\documentclass[journal]{IEEEtran}

\usepackage[utf8]{inputenc} 
\usepackage[T1]{fontenc}    
\usepackage{url}            
\usepackage{cite}
\usepackage{booktabs}       
\usepackage{amsfonts}       
\usepackage{nicefrac}       
\usepackage{microtype}      
\usepackage{threeparttable}
\usepackage{makeidx}
\usepackage{indentfirst}
\usepackage{amssymb}
\usepackage{mathtools}
\usepackage{amsmath}
\usepackage{algorithm}
\usepackage{algorithmic}
\usepackage{subcaption}
\usepackage{graphicx}
\usepackage{multirow}
\usepackage{makecell}
\usepackage{hhline}
\usepackage{xr}
\usepackage{xcolor}
\usepackage{lineno,hyperref}

\begin{document}
\title{NPENAS: Neural Predictor Guided Evolution for Neural Architecture Search}

\author{Chen Wei,
        Chuang Niu,
        Yiping Tang, 
        Yue Wang, 
        Haihong Hu, and
        Jimin Liang,~\IEEEmembership{Member,~IEEE,}
        \thanks{Manuscript received xxxx xx, 2020; revised xxxx xx, 2020; accepted xxxx xx, 2020. This work was supported in part by the National Natural Science Foundation of China under Grants 61976167 and U19B2030, and in part by the Xi'an Science and Technology Program under Grant 201809170CX11JC12. {\sl (Corresponding author: Jimin Liang.)}}
        \thanks{Chen Wei is with the School of Electronic Engineering, Xidian University, Xi’an, Shaanxi 710071, China, and also with the College of Economics and Management, Xi'an University of Posts\&Telecommunications, Xi'an, Shaanxi 710061, China (e-mail: weichen\_3@stu.xidian.edu.cn).}
        \thanks{Chuang Niu is with the Department of Biomedical Engineering, Rensselaer Polytechnic Institute, Troy, NY 12180, USA (e-mail: niuc@rpi.edu)}
        \thanks{Yiping Tang, Yue Wang, Haihong Hu and Jimin Liang are with School of Electronic Engineering, Xidian University, Xi’an, Shaanxi 710071, China (e-mail: tangyiping@aliyun.com; wangyue1991@stu.xidian.edu.cn; hhhu@mail.xidian.edu.cn; jimleung@mail.xidian.edu.cn).}
}


\IEEEtitleabstractindextext{%
\begin{abstract}
Neural architecture search (NAS) is a promising method for automatically design neural architectures. NAS adopts a search strategy to explore the predefined search space to find outstanding performance architecture with the minimum searching costs. Bayesian optimization and evolutionary algorithms are two commonly used search strategies, but they suffer from computationally expensive, challenge to implement or inefficient exploration ability. In this paper, we propose a neural predictor guided evolutionary algorithm to enhance the exploration ability of EA for NAS (NPENAS) and design two kinds of neural predictors. The first predictor is defined from Bayesian optimization and we propose a graph-based uncertainty estimation network as a surrogate model that is easy to implement and computationally efficient. The second predictor is a graph-based neural network that directly outputs the performance prediction of the input neural architecture. The NPENAS using the two neural predictors are denoted as NPENAS-BO and NPENAS-NP respectively. In addition, we introduce a new random architecture sampling method to overcome the drawbacks of the existing sampling method. Extensive experiments demonstrate the superiority of NPENAS. Quantitative results on three NAS search spaces indicate that both NPENAS-BO and NPENAS-NP outperform most existing NAS algorithms, with NPENAS-BO achieving state-of-the-art performance on NASBench-201 and NPENAS-NP on NASBench-101 and DARTS, respectively.

\end{abstract}

}

\maketitle

\IEEEdisplaynontitleabstractindextext

\IEEEpeerreviewmaketitle

\section{Introduction}
\IEEEPARstart{N}{eural} architecture search (NAS) aims to automatically design network architecture, which is essentially an optimization problem of finding an architecture with the best performance in specific search space with constrained resources \cite{Ren2020ACS, Ying2019NASBench101TR}. As the search space is often huge, a critical challenge of NAS is how to effectively and efficiently explore the search space. Among the various search strategies proposed previously, Bayesian optimization(BO) \cite{kandasamy2018neural,White2019BANANASBO} and evolutionary algorithms (EA) \cite{Real2017LargeScaleEO,Real2018RegularizedEF,9075201,8742788,8712430} are two commonly used methods. However, the previous BO- and EA-based methods for NAS are still suffering from high computational cost, challenge to implement, or inefficient exploration ability. To tackle these problems, we propose a neural predictor guided evolutionary search strategy, which fits both BO- and EA-based frameworks. Our insight comes from the following observation.

\begin{figure}[ht!]
    \centering
    \includegraphics[width=0.48\textwidth, height=3cm]{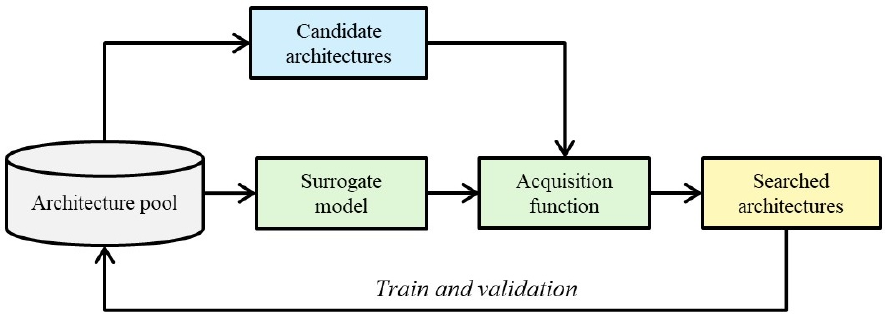}
    \caption{Pipeline of Bayesian optimization for NAS.}
    \label{fig:bo_nas}
\end{figure}

BO-based search strategy regards NAS as a noisy black-box function optimization problem. The diagram of BO for NAS is demonstrated in Fig. \ref{fig:bo_nas}. It employs a surrogate model to represent the distribution of the black-box function and utilizes an acquisition function to rank the performance of candidate architectures, which is beneficial to balance the exploration and exploitation for architecture search. Because of the huge search space, the acquisition function can't rank all the architectures. Therefore, evolutionary algorithms are usually employed to generate candidate architectures for the acquisition function to select potential superior architectures \cite{kandasamy2018neural,White2019BANANASBO}. However, previous studies adopted the Gaussian process \cite{kandasamy2018neural} or an ensemble of neural networks \cite{White2019BANANASBO} as the surrogate model, which makes the surrogate model computationally intensive and cannot be trained end-to-end.

EA-based search strategy selects several parents from a population of architectures and generates offspring through mutation from the selected parents. Because the architectures in search space exhibit locality \cite{Ying2019NASBench101TR} and EA usually generates offspring that are close to their parents \cite{Kingma2014AdamAM}, it restricts EA to efficiently explore the search space. Moreover, since previous studies using EA for NAS mostly evaluate the performance of offspring via the training and validation procedures \cite{Real2017LargeScaleEO, Real2018RegularizedEF, 9075201, 8742788, 8712430}, the searching cost is catastrophic for ordinary researchers. 

The pipeline in Fig. \ref{fig:bo_nas} can be understood from the perspective of EA, where an evolutionary algorithm selects a number of parents from the architecture pool and generates a collection of candidate architectures. Then the acquisition function is used as a performance predictor to rank the candidate architectures. Finally, the top performance architectures are selected and trained, to update the architecture pool. 

\begin{figure*}[ht!]
    \centering
    \includegraphics[width=0.98\textwidth, height=14cm]{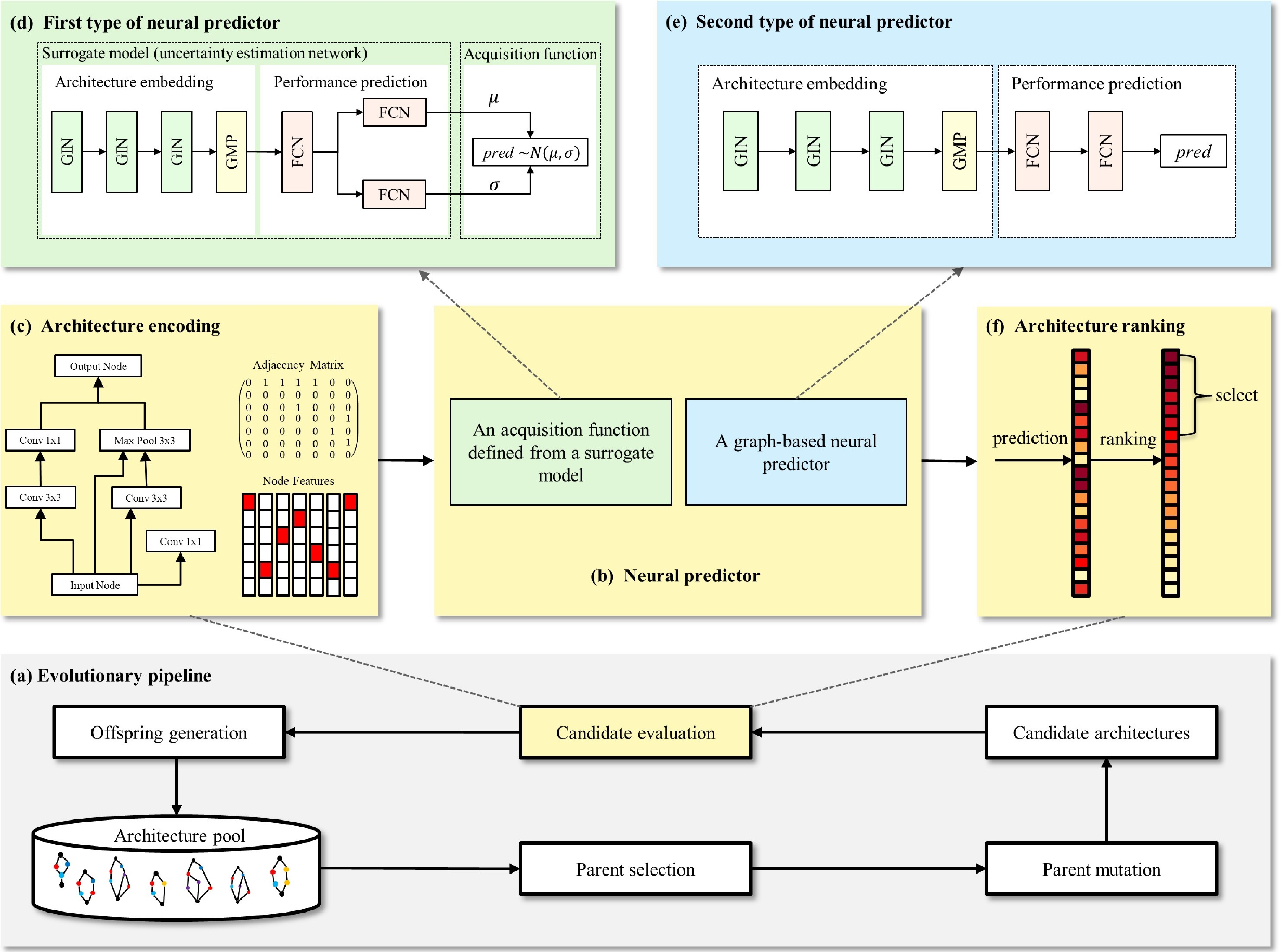}
    \caption{Overview of neural predictor guided evolutionary algorithm for NAS. (a) Pipeline of evolutionary algorithm. (b) Two kinds of neural predictors. (c) Graph encoding of neural architecture. (d) First type of neural predictor: an acquisition function defined from a graph-based uncertainty estimation network. (e) Second type of neural predictor: a graph-based neural predictor. (f) Candidate neural architecture ranking.}
    \label{fig:npenas_ea}
\end{figure*}

Inspired by the above observation, we propose a neural predictor guided evolutionary algorithm that combines EA with a neural performance predictor for NAS, named as NPENAS. The pipeline of NPENAS is illustrated in Fig. \ref{fig:npenas_ea}a. One of the main differences between NPENAS and the existing EA-based methods \cite{Real2017LargeScaleEO, Real2018RegularizedEF, 9075201, 8742788, 8712430} is that we generate multiple, instead of one, candidate architectures from each parent. This candidate generation strategy will improve the exploration ability of EA. But it will cause another problem - it is computationally expensive to evaluate all the candidates via the training and validation procedures. Therefore, we propose to utilize a neural predictor to rank the candidate architectures, and only the top performance neural architectures are selected to evaluate and update the architecture pool.

The critical issue of NPENAS is how to design neural performance predictors. In this paper, we propose two kinds of neural predictors (Fig. \ref{fig:npenas_ea}b). Firstly, since the acquisition function of BO can be viewed as a neural predictor, which needs a surrogate model to describe the distribution of the black-box function, we design a new surrogate model for BO. Specifically, we employ a graph encoding method to represent the neural architecture (Fig. \ref{fig:npenas_ea}c). Then we design a graph-based uncertainty estimation network that takes the graph encoding as input and outputs the mean and standard deviation of a Gaussian distribution to represent the distribution of the input architecture's performance (Fig. \ref{fig:npenas_ea}d). The graph-based uncertainty estimation network is adopted as the surrogate model for BO, from which an acquisition function is defined to search the potential superior architectures. We name NPENAS using this neural predictor as NPENAS-BO. The competitive advantage of the proposed surrogate model lies in its low computational complexity by avoiding the computationally intensive matrix inversion operation and its simplicity of being able to train end-to-end. 

The design of the acquisition function to effectively balance the exploration and exploitation for NAS is a cumbersome task. Therefore, instead of employing the BO theory, we propose a graph-based neural predictor to directly outputs the performance prediction of input architecture (Fig. \ref{fig:npenas_ea}e). The corresponding NPENAS method with this neural predictor is termed as NPEANS-NP. We conducted extensive experiments to demonstrate the superiority of NPENAS in comparison with existing NAS algorithms. 

Our main contributions can be summarized as follows.
\begin{itemize}
    \item We propose a neural predictor guided evolutionary algorithm for NAS, namely NPENAS. The combination of EA with a neural predictor can enhance the exploration ability of EA and is beneficial to balance the exploration and exploitation for NAS. To the best of our knowledge, this is the first paper focusing on the combination of EA with a neural predictor for NAS.
    \item We design two kinds of neural predictors for NPENAS. The first one is an acquisition function defined from a graph-based uncertainty estimation network. The second one is a graph-based neural predictor. The variants of NPENAS using these two kinds of neural predictors are termed as NPEANS-BO and NPEANS-NP respectively.
    \item We investigate the drawbacks of the default architecture sampling method on NASBench-101 \cite{Ying2019NASBench101TR} and demonstrate that sampling architectures directly from the search space is beneficial for performance improvement.
    \item We evaluate NPENAS-BO for 600 trials on the NASBench-101 benchmark \cite{Ying2019NASBench101TR}. With a search budget of 150 queried architectures, NPENAS-BO achieves a mean test error of $5.9 \%$, which is slightly better than that of BANANAS \cite{White2019BANANASBO} ($5.91 \%$). On the NASBench-201 benchmark \cite{Dong2020NASBench201ET}, with only 100 queried architectures, averaged over 600 trials NPENAS-BO achieves a mean test error of $8.93 \%$ which is close to the \textit{ORACLE} baseline \cite{Wen2019NeuralPF} $8.92 \% $. On the DARTS \cite{Liu2018DARTSDA} search space, with 2.5 GPU days, NPENAS-BO finds an architecture with the mean test error of $2.64 \%$ and the best test error of $2.52 \%$. The search speed is 4.7x faster than BANANAS \cite{White2019BANANASBO}, and the searched architecture's test error is better than that of BANANAS \cite{White2019BANANASBO} ($2.54 \%$). NPENAS-BO outperforms most of the existing NAS algorithms and is much faster than existing NAS algorithms that do not use weight sharing.
    \item NPENAS-NP achieves a mean test error of $5.86 \%$ average over 600 trials on the NASBench-101 benchmark with a search budget of 150 neural architecture queries, which is close to the \textit{ORACLE} baseline \cite{Wen2019NeuralPF} $5.77 \%$. On the NASBench-201 benchmark, with 100 queried architectures and averaged over 600 trials, NPENAS-NP achieves a mean test error of $8.95 \% $ that is comparable with the best performance $8.92 \%$. On the DARTS search space, NPENAS-NP finds an architecture that achieves state-of-the-art performance with the mean test error of $2.54 \%$ and the best test error of $2.44 \% $, with only 1.8 GPU days. The search speed is $6.5$x faster than that of BANANAS \cite{White2019BANANASBO} and is comparable with that of some gradient-based NAS algorithms, e.g., DARTS \cite{Liu2018DARTSDA} ($1.5$ GPU days).
\end{itemize}

\section{Related Works}

\subsection{Neural Architecture Encoding and Embedding} \label{na_encoding}
Neural networks are usually composed by stacking several convolution layers, fully connected layers, and other layers. The neural architecture has to be encoded into some form to be used by the search strategy. The encoding methods roughly fall into two categories - vector encoding \cite{White2019BANANASBO,9075201,8742788,8712430,Wang2019AlphaXEN} and graph encoding \cite{kandasamy2018neural,Wen2019NeuralPF,DBLP:journals/corr/abs-2004-01899}. 

The adjacency matrix encoding is the most commonly used vector encoding method \cite{Wang2019AlphaXEN, Zhou2019BayesNASAB, DBLP:conf/iclr/BakerGRN18}. Sun \textit{et al}. \cite{9075201, 8742788, 8712430} represented each layer of a neural network with a predefined vector containing the layer type, channel number, and kernel size, and then connected the layer vectors to form the final vector. BANANAS \cite{White2019BANANASBO} adopted a path-based encoding method which extracted all the input-to-output paths in neural network and converted them into a vector. However, path-based encoding is not suitable for macro-level search space because its encoding vector will increase exponentially \cite{White2019BANANASBO}. After architecture encoding, a multiple layer fully connected network \cite{White2019BANANASBO,Wang2019AlphaXEN} or recurrent neural network (RNN) \cite{zoph2016neural} is usually adopted to embed the vector encoding into feature space.

As a neural network can be defined as a direct acyclic graph (DAG), it is straightforward to represent its architecture with a graph structure. The graph nodes represent the network layers and an adjacency matrix is utilized to represent the layer connections \cite{kandasamy2018neural, Wen2019NeuralPF, DBLP:journals/corr/abs-2004-01899}. Most of the existing methods adopted a spectral-based graph neural network (GCN) \cite{Kipf2016SemiSupervisedCW} to generate feature embedding. But GCN is designed to process undirected graph, while neural network architecture is directed graph. 

Following previous methods, we also adopt DAG to define the neural architectures in this paper. Unlike existing methods, we define a new node type to represent isolated node and convert the original graph into a new graph that does not contain any isolated node, as graph neural network is not suitable to process graphs with isolated node \cite{DBLP:journals/spm/ShumanNFOV13,DBLP:conf/icml/WuSZFYW19}. None of the previous methods for NAS considered this factor. Furthermore, we adopt a spatial-based graph neural network graph isomorphism network (GIN) \cite{Xu2018HowPA} to generate feature embedding as it can process directed graph neural network from a message passing perspective \cite{Gilmer2017NeuralMP}. 
Empirical experiments demonstrate that the proposed neural architecture encoding and embedding method are efficient and straightforward.

\subsection{Search Strategy for NAS}
Search strategy plays a critical role in efficiently exploring the search space for NAS. The commonly used search strategies include reinforcement learning (RL) \cite{zoph2016neural,enas,Zoph2017LearningTA,Liu2017ProgressiveNA},  gradient-based methods \cite{Liu2018DARTSDA,Zhou2019BayesNASAB,Xie2019SNASSN,Chen2019ProgressiveDA,xu2020pcdarts}, Bayesian optimization \cite{kandasamy2018neural,White2019BANANASBO}, evolutionary algorithms, \cite{Real2017LargeScaleEO,Real2018RegularizedEF,9075201,8742788,8712430} and predictor-based methods \cite{Liu2017ProgressiveNA,Wen2019NeuralPF,Wang2019AlphaXEN,DBLP:journals/corr/abs-2004-01899}. 

RL for NAS takes a recurrent neural network (RNN) as a controller to generate neural architectures sequentially. As NAS is essentially an optimization problem and RL is a more difficult problem than optimization \cite{kandasamy2018neural}, the RL search strategy always needs thousands of GPU days to train the controller. 

Gradient-based methods adopt a single super-network to represent NAS's search space and architecture in the search space can be realized by sampling a sub-network via an architecture distribution from the super-network. All the sampled architectures share weights in the same super-network, which can speed up the evaluation of each sampled architecture. This kind of method can also be categorized as weight sharing method from the perspective of architecture evaluation. Some previous research has reported that the gradient-based methods typically incur a large bias and suffer from instability \cite{Elsken2018NeuralAS,Li2019RandomSA,Sciuto2019EvaluatingTS}. 

Bayesian optimization utilizes an acquisition function defined from a surrogate model to sample the potential optimal solution\cite{Shahriari2016TakingTH, Frazier2018ATO}. NASBOT \cite{kandasamy2018neural} adopted a Gaussian process as the surrogate model and presented a distance metric to calculate the kernel function, which involved a computationally intensive matrix inversion operation. BANANAS \cite{White2019BANANASBO} employed a collection of identical meta neural networks to predict the accuracy of candidate architectures, which avoided the inverse matrix operation. BANANAS outperformed a variety of NAS methods and achieve state-of-the-art error on NASBench-101 \cite{Ying2019NASBench101TR}. However, the ensemble of meta neural networks prohibits the end-to-end training of neural predictors. Due to the superior performance of BANANAS, we choose BANANAS as the baseline for comparison in this paper.

Previous EA-based methods \cite{Real2017LargeScaleEO,Real2018RegularizedEF,9075201,8742788,8712430} for NAS mainly focused on the design of architecture encoding, parent selection, and mutation strategies. All of them adopted one-to-one mutation operation, i.e., one parent generates one offspring. Due to the locality of the search space, a property by which the architectures that are ``close by'' tend to have similar performance metrics \cite{Ying2019NASBench101TR}, and the phenomenon that EA usually generates offspring close to their parents \cite{Kingma2014AdamAM}, the searching overheads of EA are always enormous.

Predictor-based NAS utilizes an approximated performance predictor to find promising architectures for further evaluating. Liu \textit{et al}. \cite{Liu2017ProgressiveNA} designed a neural predictor to guide the heuristic search to search the space of cell structures, starting with shallow models and progressing to complex ones.  AlphaX \cite{Wang2019AlphaXEN} explored the search space via Monte Carlo Tree Search (MCTS) and proposed a Meta-Deep Neural Network as a neural predictor to speed up the exploration. Wen \textit{et al}. \cite{Wen2019NeuralPF} adopted a complicated cascade neural predictor to rank all the architectures in search space, select the top-k architectures to fully evaluate, the architecture with the best performance from the evaluated architectures is reported. Ning \textit{et al}. \cite{DBLP:journals/corr/abs-2004-01899} proposed a graph-based neural architecture encoding scheme as a neural predictor and illustrated its performance by utilizing the neural predictor to integrate with existing NAS methods. 

In this paper, we propose to adopt the one-to-many mutation strategy, i.e., one parent generates multiple candidate architectures, to enhance the exploration ability of EA for NAS. In order to alleviate the computational burden of candidate architectures evaluation, we propose to utilize a neural predictor to rank the candidate architectures. Our study focuses on the design of the neural predictor and its integration with EA for NAS. Because of our proposed neural predictors' superior performance, that may be of independent interest beyond the combining with EA for NAS.

\section{Methodology}
In order to enhance the exploration ability of EA for NAS, we adopt a one-to-many mutation strategy and a neural predictor to guide the evolution of EA. The pipeline of the evolutionary algorithm is shown in Fig. \ref{fig:npenas_ea}a. The encoding method of neural architecture is introduced in Section \ref{architecture_encoding_method}. The proposed neural predictors are expounded in Section \ref{neural_predictor}. The combination of the evolutionary algorithm with the graph-based uncertainty estimation network is outlined in Section \ref{npubo}, and the combination of the evolutionary algorithm with graph-based neural predictor is discussed in Section \ref{npenas}. Moreover, random architecture sampling methods are analyzed in Section \ref{random_architecture_sample}.

\subsection{Problem Formulation}
NAS can be modeled as a global optimization problem. Given a search space $S$, the goal of NAS can be formulated as 
\begin{equation} \label{eq:1}
    s^* = \arg \mathop{\min}_{s \in S} f(s),
\end{equation}
where $f$ is a performance measurement of the neural architecture $s$. 
As it is impossible to evaluate the performance of all the architectures via the training-and-validation procedure within the limited searching cost, some kind of search strategy is always demanded by NAS.

\subsection{Neural Architecture Encoding} \label{architecture_encoding_method}
An neural architecture $s$ can be described as a DAG 
\begin{equation} \label{eq:DAG}
    s=(V, E), 
\end{equation}
where $V=\{v_i\}_{i=1:N}$ is the set of nodes representing layers in the neural architecture, and $E=\{v_i, v_j\}_{i,j=1:M}$ is the set of edges describing the connections between layers. As we will compare the proposed NPENAS with other algorithms on three different search spaces, i.e., NASBench-101 \cite{Ying2019NASBench101TR}, NASBench-201, \cite{Dong2020NASBench201ET} and DARTS \cite{Liu2018DARTSDA}, we need to encode the neural architecture on these search spaces respectively. 

\paragraph{NASBench-101} NASBench-101 designs a cell level search space. Each cell is composed of one input node, one output node, and at most five operation nodes. The operation node has three types, i.e., conv 1$\times$1, conv 3$\times$3, maxpool 3$\times$3. We define a new node type "isolated". If an isolated node appears in a cell, it will be connected with the input node and its node type will be marked as isolated. The added connection will eliminate the isolated node in the cell. Since the isolated node is not connected to the output node, this added connection will not affect the performance of neural architecture. If the number of nodes in the cell is less than seven, we ensure that all cells in the search space have seven nodes by inserting isolated nodes.

As there are seven nodes in each cell and six node types in the search space, we adopt a $7 \times 7$ upper triangle adjacency matrix to represent $E$ and use a $6$-dimensional one-hot vector as the node feature. An example of the neural network cell in NASBench-101 is illustrated in the left of Fig. \ref{fig:npenas_ea}c, with the corresponding adjacency matrix and node features shown in the right.

\begin{figure}[ht!]
    \begin{center}
        \begin{subfigure}{0.24\textwidth}
            \includegraphics[width=0.98\linewidth, height=1.5cm]{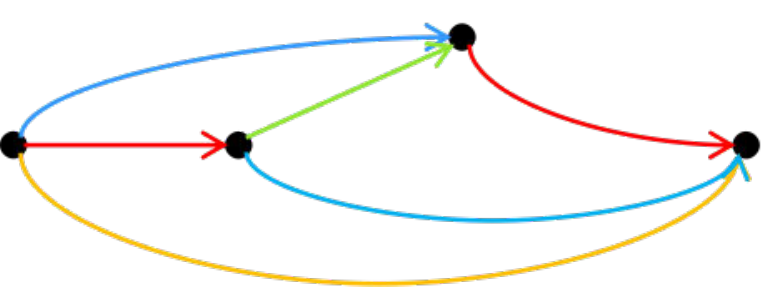}
            \caption{Original graph representation of cell in NASBench-201.}
            \label{nas_bench_201_subfig:1}
        \end{subfigure}
        \begin{subfigure}{0.24\textwidth}
            \includegraphics[width=0.98\linewidth, height=1.5cm]{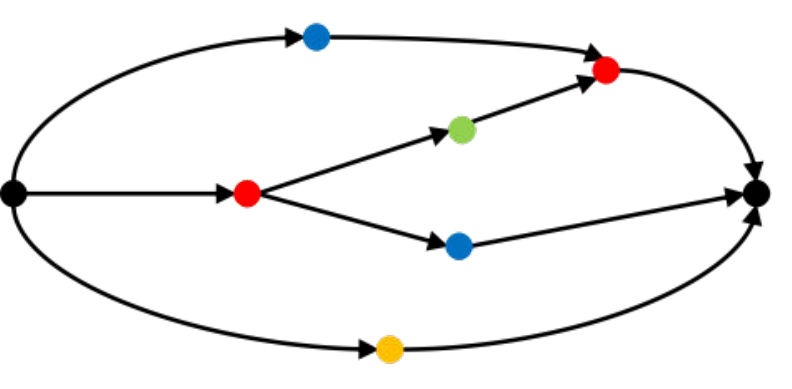}
            \caption{Converted graph representation of (a).}
            \label{nas_bench_201_subfig:2}
        \end{subfigure}
    \end{center}
    \caption{Graph representation of cell in NASBench-201.}
    \label{nas_bench_201_DAG}
\end{figure}

\paragraph{NASBench-201} Similar to NASBench-101, the NASBench-201 is a cell level search space, where each cell is defined via a DAG. A cell in NASBench-201 uses four nodes to represent the sum of feature maps and edges to represent operations, as illustrated in Fig. \ref{nas_bench_201_subfig:1}. This scheme is inconsistent with the above DAG representation in Eq \ref{eq:DAG}, so we transform the original cell representation in NASBench-201 to a new graph where the nodes represent operations (layers) and the edges stand for layer connections, as shown in Fig. \ref{nas_bench_201_DAG}. There are eight nodes in each cell and eight node types (operations) in the search space after conversion, therefore, we adopt an $8 \times 8$ upper triangle adjacency matrix to represent the layer connection and a collection of $8$-dimensional one-hot encoded vector as nodes feature. 

\paragraph{DARTS} DARTS search space defines two types of neural network cells - the normal cell and the reduction cell. The cells are represented by DAG in the same manner as in NASBench-201. We use the same method as in NASBench-201 to convert the cell graph representation into the form consistent with the predefined $V$ and $E$ in Eq \ref{eq:DAG}. Then, we adopt a $15 \times 15$ upper triangle adjacency matrix and a collection of $11$-dimensional one-hot encode vectors to encode the cells. The macro neural network in DARTS is composed by sequentially stacking several normal cells and inserting the reduction cells in specific positions between the normal cells. In order to characterize the sequential connection of normal and reduction cells, we build a $30 \times 30$ adjacency matrix, and assign the upper-left $15\times15$ elements and the bottom-right $15\times15$ elements with the adjacency matrix of normal cells and reduction cells, respectively. The input of the reduction cell is set as the output of the normal cell to describe the sequential connection. Finally, a cell of DARTS is encoded by the $30\times30$ adjacency matrix together with the node features of the normal and reduction cells.

\subsection{Neural Predictor} \label{neural_predictor}
We design two kinds of neural predictors for neural architecture evaluation: 1) an acquisition function defined from a graph-based uncertainty estimation network, and 2) a graph-based neural predictor. 

The first kind of neural predictor is inspired by the existing BO-based methods for NAS. Bayesian optimization usually utilizes the Gaussian process as surrogate model to describe the prior confidence of the performance prediction function $f$, which can be formulated as
\begin{equation} \label{eq:2}
    f(s) \thicksim GP(\mu(s),k(s,s^\prime)), \quad s,s^\prime \in S,
\end{equation}
where $\mu(s)$ is the mean function and $k(s,s^\prime)$ the kernel function. 
The calculation of kernel function $k$ requires a distance measure and involves the computational intensive operation of matrix inversion. 

We assume that the neural architectures in search space are independent and identically distributed. In this case, the distribution of performance prediction function $f$ can be fully defined by its mean function $\mu(s)$ and standard deviation function $\sigma(s)$,  

\begin{equation} \label{eq:3}
    f(s) \thicksim N(\mu(s),\sigma(s)), \quad s \in S.
\end{equation}

In this paper, a graph-based uncertainty estimation network is designed to implement the mean and standard deviation functions in Eq \ref{eq:3} and is used as the surrogate model to describe the distribution of performance prediction function $f$. The uncertainty estimation network consists of two parts - architecture embedding and performance prediction, as shown in Fig. \ref{fig:npenas_ea}d. The architecture embedding part is implemented by three sequentially connected spatial graph neural network GINs \cite{Xu2018HowPA} followed by a global mean pooling (GMP) layer. Each GIN uses multi-layer perceptrons (MLPs) to update each node by aggregating features of its neighbors. For neural performance prediction, the surrogate model passes the embedded feature through several fully connected layers and outputs the estimation of the mean $\mu$ and standard deviation $\sigma$ of the Gaussian distribution that describes the confidence of the input architecture's performance. Thereafter, the \textit{Thompson Sampling} (TS) \cite{Russo2017ATO} is adopted as the acquisition function defined from our proposed surrogate model to generate each architecture's performance. 

The second kind of neural predictor is a graph-based neural network that also consists of two functional parts, as shown in Fig. \ref{fig:npenas_ea}e. The architecture embedding part is identical to the above surrogate model. The performance prediction is implemented by passing the embedded feature through several fully connected layers and a sigmoid layer. 

\subsection{Neural Predictor Guided Evolutionary Algorithm for NAS (NPENAS)}
The pipeline of the proposed NPENAS is shown in Fig. \ref{fig:npenas_ea}a. Firstly, an architecture pool is initialized with randomly sampled architectures that are evaluated by training and validation. The architectures in the pool are selected sequentially as parents for mutation. A one-to-many mutation strategy is adopted to generate a predefined number of non-isomorphism candidate architectures. Then a neural predictor is used to rank the candidate architectures. The top performance architectures are selected as offspring and evaluated by training and validation. Finally, the offspring are appended to the architecture pool. The procedure repeats a given number of times, and the neural predictor is trained from scratch with all the architectures in the pool at each iteration. 

As we present two kinds of neural predictors, the corresponding implementations of NPENAS, namely NPENAS-BO and NPENAS-NP, are detailed as follows. 

\subsubsection{NPENAS-BO} \label{npubo}

The \textit{maximum likelihood estimate} (MLE) is used as loss function to optimize the graph-based uncertainty estimation network(denoted as $G_u$) with parameter $w_u$. It can be formulated as 

\begin{equation} \label{eq:6}
    \begin{aligned}
        w_u^*  = \arg \max_{w_u} \prod_{(s_i,y_i) \in D} P(y_i|\mu_i, \sigma_i), \quad (\mu_i, \sigma_i) = G_u(s_i),
    \end{aligned} 
\end{equation}
where $D=\{(s_i, y_i), i=1,2,\cdots,n\}$ is the training dataset, i.e., the architecture pool, $n$ is the number of architectures in the pool, $s_i$ is a neural architecture and $y_i$ its validation error.

The procedure of NPENAS-BO is summarized in Algorithm .\ref{algo:1}.

\begin{algorithm}[ht!]
\caption{NPENAS-BO}
\label{algo:1}
\textbf{Input:} Search space $S$, initial population size $n_0$, initial population $D=\{(s_i, y_i), i=1,2,\cdots,n_0\}$, neural predictor $G_u$, number of total training samples $total\_num$, number of candidate architectures $mu\_num$ and number of offspring $t$.\\
\textbf{For} $n$ from $n_0$ \textbf{to} $total\_num$ \textbf{do}
\begin{enumerate}
    \item Randomly initialize the parameters of neural predictor $G_u$;
    \item Train the neural predictor $G_u$ with dataset $D=\{(s_i,y_i), i=1,2,\cdots,n\}$;
    \item Sequentially select the architectures in $D$ based on their validation error and utilize the one-to-many mutation strategy to generate $mu\_num$ candidate architectures $M=\{s_m, m=1,2,\cdots,mu\_num\}$;
    \item Use $G_u$ to predict the mean $\mu$ and standard deviation $\sigma$ of each candidate architecture in $M$;
    \item Use acquisition function TS to predict the performance of  architectures in $M$, $f(s_m) \thicksim N(G_u(s_m)), m=1,2,\cdots,mu\_num$;
    \item Based on $f(s_m)$, select the top $t$ architectures from $M$ as the offspring $M_t=\{s_m, m=1,2,\cdots,t\}$;
    \item Train the architectures in $M_t$ to obtain their validation errors $y_m, m=1,2,\cdots,t$;
    \item Append $M_t$ to $D$;
    \item $n:=n+t$;
\end{enumerate}
\textbf{End For} \\
\textbf{Output:} $s^* = \arg \min (y_i), {(s_i,y_i)\in D}$.
\end{algorithm}

\subsubsection{NPENAS-NP} \label{npenas}
The procedure of NPENAS-NP is summarized in Algorithm. \ref{algo:2}, where the graph-based neural predictor, denoted as $G$ with parameter $w$, is optimized using the \textit{mean square error} (MSE) loss function. It is formulated as

\begin{equation}\label{eqn_ver2:1}
    w^* = \arg\min_w \sum_{(s_i, y_i) \in D} (G(s_i) - y_i)^2.
\end{equation}

\begin{algorithm}[ht!]
    \caption{NPENAS-NP}
    \label{algo:2}
    \textbf{Input:} Search space $S$, initial population size $n_0$, initial population $D=\{(s_i, y_i), i=1,2,\cdots,n_0\}$, neural predictor $G$, number of total training samples $total\_num$, number of candidate architectures $mu\_num$ and number of offspring $t$.\\
    \textbf{For} $n$ from $n_0$ \textbf{to} $total\_num$ \textbf{do}
    \begin{enumerate}
        \item Randomly initialize the parameter of neural predictor $G$;
        \item Train the neural predictor $G$ with dataset $D=\{(s_i,y_i), i=1,2,\cdots,n\}$;
        \item Sequentially select architectures in $D$ based on their validation error and utilize the one-to-many mutation strategy to generate $mu\_num$ candidate architectures $M=\{(s_m), m=1,2,\cdots,mu\_num\}$;
        \item Use $G$ to predict the performance of architectures in $M$, $f(s_m)=G(s_m), m=1,2,\cdots,mu\_num$;
        \item Based on $f(s_m)$, select the top $t$ architectures from $M$ as offspring $M_t=\{(s_m), m=1,2,\cdots,t\}$;
        \item Train the architectures in $M_t$ to obtain their validation errors $y_m, m=1,2,\cdots,t$;
        \item Append $M_t$ to $D$;
        \item $n:=n+t$;
    \end{enumerate}
    \textbf{End For}\\
    \textbf{Output:} $s^* = \arg \min (y_i), {(s_i,y_i)\in D}$.
\end{algorithm}

\subsection{Random Architecture Sampling} \label{random_architecture_sample}
The first step of NPEANS is to initialize the architecture pool by randomly sampling the search space. NASBench-101 \cite{Ying2019NASBench101TR} provides a default sampling method. Firstly, a random graph $s_r$, represented by a $7 \times 7$ adjacency matrix and its corresponding node operations, is generated. The extraneous parts of $s_r$, if exists, will be pruned and a new graph $s_{rp}$ will be generated. If $s_{rp}$ meets all the requirements defined by NASBench-101, it will be used to query its evaluation metrics from NASBench-101. Finally, $s_r$ is selected as the randomly sampled neural architecture. The sampling procedure repeats until a given number of architectures are generated.

The default sampling pipeline has two adverse effects on NAS. Firstly, the extraneous part pruning operation can easily lead to generating multiple architectures with the same evaluation metrics. The same problem occurs in the mutation procedure of EA. This will make it more difficult to train the predictor for NAS. 

Secondly, we find out experimentally that the default sampling method tends to generate architectures in a subspace of the real underlying search space. This finding is obtained by utilizing the path-based encoding method \cite{White2019BANANASBO} to investigate the distribution of sampled architectures. Path-based encoding is a vector-based architecture encoding method, and it characterizes an architecture by its input-to-output paths. As there are 364 distinct paths in the NASBench-101 search space, an architecture can be represented by a $364$-dimensional binary vector $\mathbf{c}=(c_1, c_2, \cdots, c_{364})$. 
We propose to represent the sampled architecture distribution by path distribution, i.e., the frequency of path occurrence in the sampled architectures. 
Denote the set of path-based encoding vectors of $n$ architectures as 
$ C=\{\mathbf{c_1}, \mathbf{c_2}, \cdots, \mathbf{c_n}\}$, its path distribution is defined as 
\begin{equation} \label{eq:hist_dist}
    \mathbf{p} = \log \left( \frac{1}{T} \sum_{\mathbf{c_i} \in C} \mathbf{c_i} \right), \quad where \quad T = \sum_{\mathbf{c_i} \in C} \sum_{j=1}^{364} c_{ij}, 
\end{equation}
where $c_{ij}$ is the $j^{th}$ element of vector $\mathbf{c_i}$.

\begin{figure}[ht!]
    \begin{center}
        \includegraphics[width=0.98\linewidth, height=6.3cm]{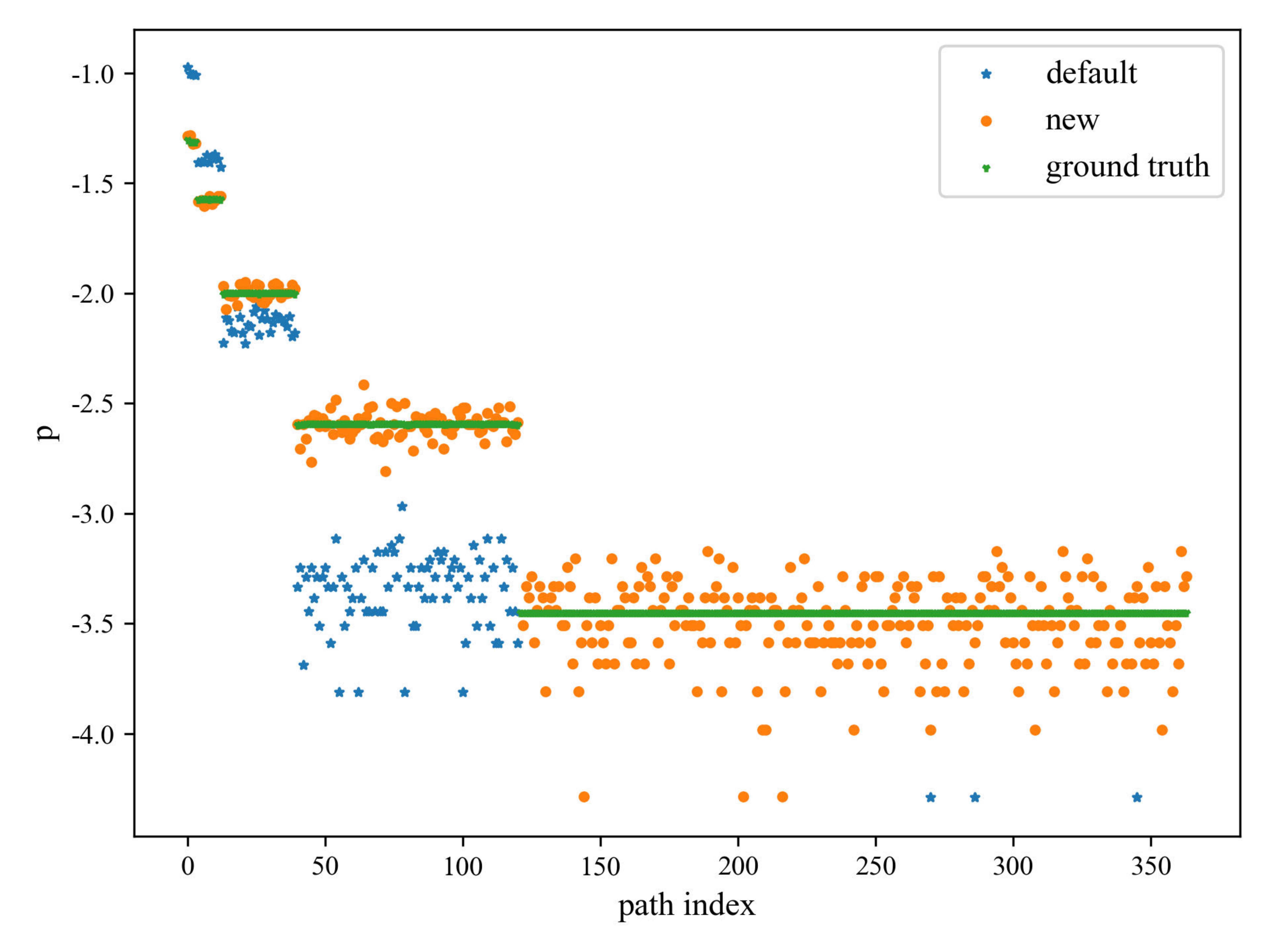}
        \caption{Path distributions of different sampling methods and the ground-truth distribution.} 
        \label{architecture_sampling_pipeline:subfig_2}
    \end{center}
\end{figure}

The path distribution of 5k architectures sampled using the default sampling method and that of all the architectures in NASBench-101 (regarded as the ground truth) are shown in Fig. \ref{architecture_sampling_pipeline:subfig_2}. We can see that the default sampling method tends to sample architectures with low path indices. 

In order to eliminate the above adverse effects, we propose a new sampling method to directly sample the architectures in search space. Specifically, a dictionary $\{ key, value\}$ is constructed with all the architectures in the search space. The adjacency matrix and node feature of the architectures are used as $value$ and their corresponding hash codes as $key$. We select a given number of keys by randomly sampling from the $keys$' list, and then get the sampled architectures by utilizing the keys to query from the dictionary. 

Using our proposed sampling method, the path distribution of 5k randomly sampled architectures is also shown in Fig. \ref{architecture_sampling_pipeline:subfig_2}. It can be seen that the path distribution from the new sampling method is consistent with the ground-truth. Concretely, the \textit{KL-divergence} \cite{DBLP:books/lib/Bishop07} between the path distribution of the new sampling method and the ground truth is much smaller than that between the default sampling method and the ground truth (0.009 vs. 0.3115). 

The benefits of the proposed sampling method for NAS are validated through comparative experiments on the NASBench-101 search space in Section \ref{experiments}. Since the same problem occurs in the NASBench-201 search space, we also use the proposed sampling method for experiments on it. The default architecture sampling method provided by DARTS is essentially consistent with the proposed sampling method and is therefore used directly in our experiments.

\section{Experiments and Analysis} \label{experiments}
This section presents the empirical performance of NPENAS-BO and NPENAS-NP. All the experiments are implemented in Pytorch \cite{Paszke2019PyTorchAI}. We use the implementation of GIN from the graph neural network library pytorch\_geometric \cite{FeyLenssen2019}. The code of NPENAS is available at \cite{Npenascode}.

\subsection{Prediction Analysis}
\paragraph{Dataset} We compare the performances of neural predictors using the NASBench-101 benchmark \cite{Ying2019NASBench101TR}. NASBench-101 is the largest benchmark dataset for NAS, which contains $423$k architectures for image classification. All the architectures are trained and evaluated multiple times on CIFAR-10 \cite{Krizhevsky09learningmultiple}. NASBench-101 defines a cell level search space, and a macro architecture containing sequentially connected normal cells and fixed reduction cells. The best architecture in the NASBench-101 search space achieves a mean test error of $5.68 \%$, and the architecture with the best mean validation error achieves a mean test error of $5.77 \%$. Since search strategies utilize the validation error of architectures to explore the search space, the mean test error of the architecture with the best validation error is a more reasonable best performance of the search space, and this best performance is denoted as the \textit{ORACLE} baseline. The \textit{ORACLE} baseline of NASBench-101 is $5.77 \%$.

\paragraph{Setup} We compare our proposed neural predictors with the meta neural network proposed by BANANAS \cite{White2019BANANASBO} under different training set sizes and two random architecture sampling methods. The mean percent error on the validation set of CIFAR-10 is used to compare their performances. Since the outputs of the graph-based uncertainty estimation network are the numerical characteristics of the distribution of the input architecture, i.e., the mean and standard deviation, we only use the mean for comparison.

Our proposed neural predictors take a graph encoding of neural architecture as input, while the meta neural network takes a vector encoding of neural architecture as input. BANANAS proposed a path-based encoding to represent neural architecture. It also compared path-based encoding with adjacency matrix encoding, which is constructed by the binary encoding of the adjacency matrix and one-hot encoding for the operations on each node. Therefore, we compare four methods in the experiments, graph-based neural predictor with graph encoding as input (NPGE), graph-based uncertainty estimation network with graph encoding as input (NPUGE), meta neural network with path-based encoding as input (MNPE), and meta neural network with adjacency matrix encoding as input (MNAE).

\paragraph{Predictor Training} 
As shown in Fig. \ref{fig:npenas_ea}d and Fig. \ref{fig:npenas_ea}e, the proposed graph-based uncertainty estimation network and the graph-based neural predictor are composed of three GIN layers followed by several fully connected layers. The graph-based uncertainty estimation network employs CELU\cite{Barron2017ContinuouslyDE} as activation function and graph-based neural predictor employs ReLU as the activation function. All GIN layers and fully connected layers utilize activation function followed by batch normalization layer\cite{Ioffe2015BatchNA}. A dropout layer is used after the first fully connected layer of the graph-based uncertainty estimation network and the graph-based neural predictor, and the dropout rate is $0.1$. The architecture details of the two networks can be found in Appendix \ref{table_detail_app_pp1} and \ref{table_detail_app_pp2}. 

All neural predictors are trained in three different training set sizes of 20, 100, and 150. The training architectures are randomly sampled from the search space. The test set size is 500, and the test architectures are also randomly sampled from the search space and do not overlap with the training set. Our proposed predictors employ Adam optimizer \cite{Kingma2014AdamAM} with initial learning rate $5\mathrm{e}{-3}$ and weight decay $1\mathrm{e}{-4}$. The graph-based uncertainty estimation network is trained $1000$ epochs with batch size $16$, and the graph-based neural predictor is trained $300$ epochs with batch size $16$. The setting of the meta neural network is the same as BANANAS \cite{White2019BANANASBO}. Unless explicitly state otherwise, we use the above setting to carry out all the experiments in the following sections.

\begin{table}
    \scriptsize
    \caption{Performance comparison of four prediction methods}
\label{table:1}
\centering
\begin{tabular}{ccccc}
    \hline
    \multicolumn{2}{c}{\multirow{2}{*}{\thead{\\Training set size}}} &
    \multicolumn{3}{c}{Testing Err Avg (\%)} \\
     \cmidrule(lr){3-5}
    &  & 20 & 100  & 150  \\ \hline
    \multirow{ 4 }{*}{ \thead{Default \\ sampling \\ method} } & MNAE & 2.951$\pm$ 0.423 & 2.626 $\pm$ 0.196 & 2.666 $\pm$ 0.200 \\
    \hhline{~~~~~}  & MNPE  & \textbf{2.318 $\pm$ 0.381} & \textbf{1.236 $\pm$ 0.178} & \textbf{1.140 $\pm$ 0.192} \\
    \hhline{~~~~~}  & NPUGE & 2.913 $\pm$ 1.134 & 1.583 $\pm$ 0.196 & 1.412 $\pm$ 0.150 \\
    \hhline{~~~~~}  & NPGE & 2.901 $\pm$ 0.750 & 1.559 $\pm$ 0.205 & 1.355 $\pm$ 0.183 \\ 
    \multicolumn{2}{c}{\textit{KL-divergence}} & 0.400 $\pm$ 0.102 & 0.152 $\pm$ 0.034 & 0.117 $\pm$ 0.025\\ \hline
     \multirow{ 4 }{*}{ \thead{New \\ sampling \\ method} } & MNAE & 2.641 $\pm$ 0.821 & 2.094 $\pm$ 0.371 & 2.121 $\pm$ 0.369 \\
    \hhline{~~~~~}  & MNPE  & 2.574 $\pm$ 0.499  & 2.304 $\pm$ 0.413 & 2.340 $\pm$ 0.398 \\
    \hhline{~~~~~}  & NPUGE & 2.884 $\pm$ 1.608 & 1.936 $\pm$ 0.372 & 1.839 $\pm$ 0.309 \\
    \hhline{~~~~~}  & NPGE  & \textbf{2.521 $\pm$ 0.636} & \textbf{1.783 $\pm$ 0.287} & \textbf{1.678 $\pm$ 0.260} \\ 
    \multicolumn{2}{c}{\textit{KL-divergence}} & 1.059 $\pm$ 0.164 & 0.481 $\pm$ 0.062 & 0.390 $\pm$ 0.049 \\ \hline
\end{tabular}
\end{table}

\paragraph{Resutls} Table \ref{table:1} presents the mean test errors of the four prediction methods with different experiment settings, averaged over 300 repeated experiments. The following observations can be made from the results. 

Firstly, with 20 training samples, all four methods have large test errors. It is a natural consequence of inadequate training samples. Therefore, we focus on the analysis of the other two sets of results. 

Secondly, except for MNAE, the new sampling method leads to performance degradation for the other three predictors. We interpret this result in terms of the \textit{KL-divergence} between the path distributions of the training and testing architectures. As shown in Table \ref{table:1}, when the training set size is 100 or 150, the average \textit{KL-divergence} of the default sampling method is three times less than that of the new sampling method. We believe this is a result of the reduced search space caused by the default sampling method. The reduced search space decreases the diversity of the training and testing architectures, thus increases the prediction performance on the testing set. While the new sampling method, albeit the test errors of the predictors are larger, maintains the architecture diversity in the underlying search space. This helps to improve the generalization ability of the neural predictor, thus improving the performance of the NAS algorithms. The experimental results in Section \ref{close_domain_results} support this interpretation. 

Thirdly, with the default sampling method, MNAE performs inferior to MNPE. This is consistent with the result in \cite{White2019BANANASBO}. However, the use of the new sampling method leads to the opposite conclusion. Since MNAE directly uses the flatten of adjacency matrix for encoding, its performance is heavily influenced by the many-to-one mapping (i.e., multiple architectures with the same evaluation metrics) phenomenon caused by the default sampling method. This finding confirms the need for the new sampling method and indicates that adjacency matrix encoding is an efficient encoding method when used properly.

Finally, in the case of using the new sampling method, our proposed graph-based neural predictor achieves the best performance, and the graph-based uncertainty estimation network also exhibits better performance than the baseline methods MNAE and MNPE. 

\subsection{Mutation Strategy Analysis} \label{mutation_strategies_exp}
The one-to-many mutation strategy is beneficial for EA to efficiently explore the search space for NAS. We compare one-to-one and one-to-many mutation strategies on the NASBench-101 benchmark to verify the above observation. The experiment adopts a standard EA pipeline. The parents are selected via binary tournament selection. One parent generates multiple candidate architectures, and the performance of each candidate architecture is obtained by training and validation. Lastly, the top-10 best candidate architectures are selected and appended to the population. At each update of the population, the test error of the best architecture is reported. 

\begin{figure}[ht!]
    \begin{center}
        \includegraphics[width=0.98\linewidth, height=6.3cm]{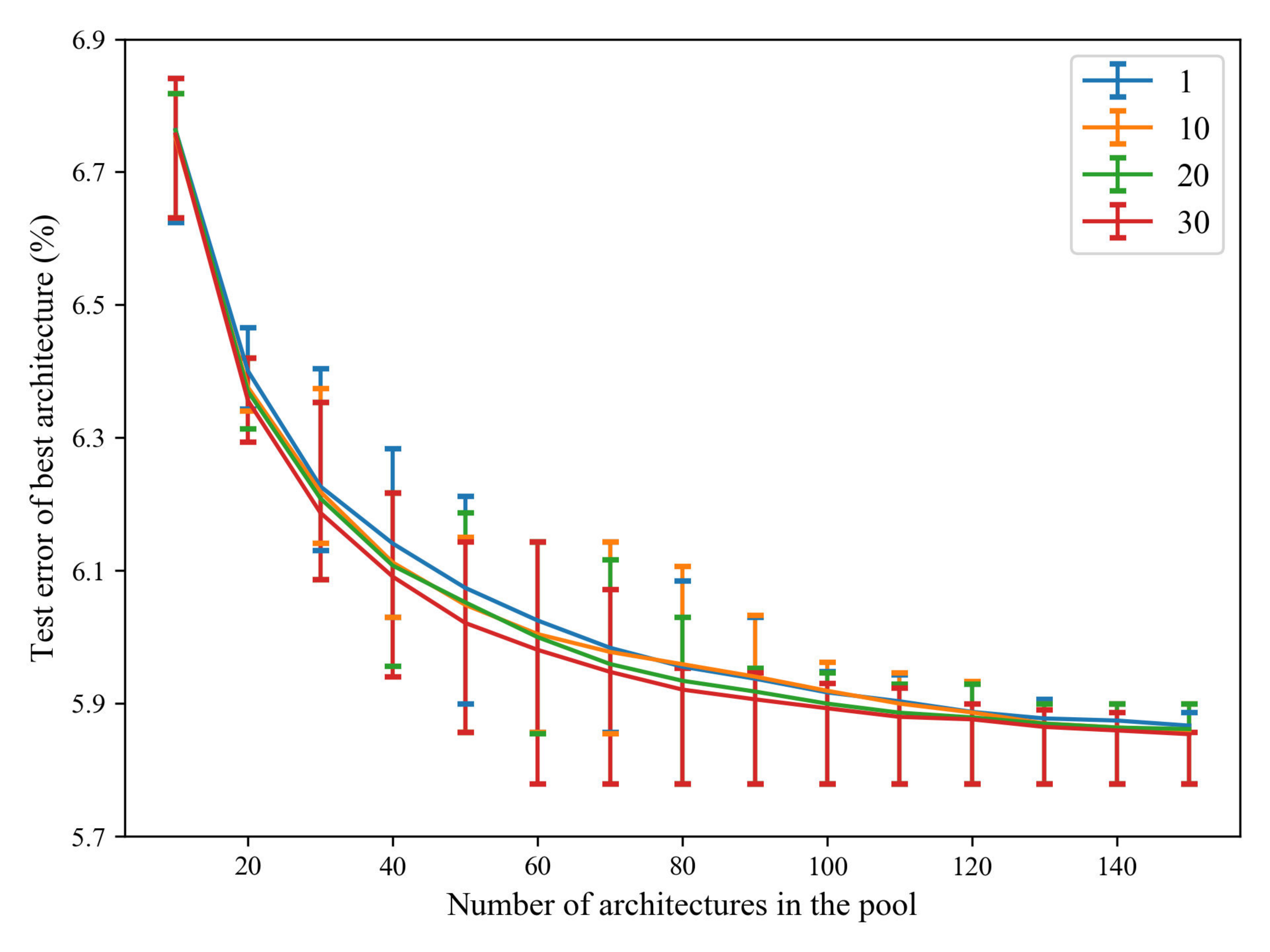}
        \caption{Comparison of different mutation strategies.}
        \label{fig:mutation_compare}
    \end{center}
\end{figure}

The number of offspring per parent is set to 1, 10, 20, and 30. The experiments are independently carried out 600 trials. The results are shown in Fig. \ref{fig:mutation_compare}.
It can be seen that the more candidate architectures a parent generate, the better the performance of the searched neural architecture. It should be noted that when the size of the architecture pool increases to a certain extent, the impact of the number of candidate architectures decreases. 

\subsection{Closed Domain Search} \label{close_domain_results}
We compare the proposed NPENAS method with the random search (RS) \cite{Li2019RandomSA}, regularized evolutionary algorithm (REA) \cite{Real2018RegularizedEF}, BANANAS \cite{White2019BANANASBO} with path-based encoding (BANANAS-PE), BANANAS with adjacency matrix encoding (BANANAS-AE), and AlphaX \cite{Wang2019AlphaXEN} on close domain search. The experiments are conducted on both the NASBench-101 \cite{Ying2019NASBench101TR} and the NASBench-201 \cite{Dong2020NASBench201ET} benchmark. NASBench-201 is a newly proposed NAS search space that contains $15.6$k architectures. All the architectures in the NASBench-201 are evaluated on three datasets - CIFAR-10 \cite{Krizhevsky09learningmultiple}, CIRAR-100 \cite{Krizhevsky09learningmultiple}, and ImageNet \cite{Krizhevsky2017ImageNetCW}. We adopt the architecture performance on CIFAR-10 for comparison. In the NASBench-201 search space, the best architecture evaluated on CIFAR-10 achieves a mean test error of $8.48\%$, and the architectures with the best validation error have a mean test error of $8.92\%$. The \textit{ORACLE} baseline of NASBench-201 evaluated on the CIFAR-10 is $8.92 \%$.

The same experiment settings as BANANAS \cite{White2019BANANASBO} are adopted. On the NASBench-101 benchmark, each algorithm is given a budget of 150 queries, and on the NASBench-201 benchmark, the budget is 100 queries. Every update of the population, each algorithm returns the architecture with the lowest validation error so far, and its corresponding test error is reported, so there are 15 or 10 best architectures in total. We run 600 trials for each algorithm and report the averaged results. 

\begin{figure}[ht!]
    \begin{center}
        \begin{subfigure}{0.48\textwidth}
            \includegraphics[width=0.98\linewidth, height=6.3cm]{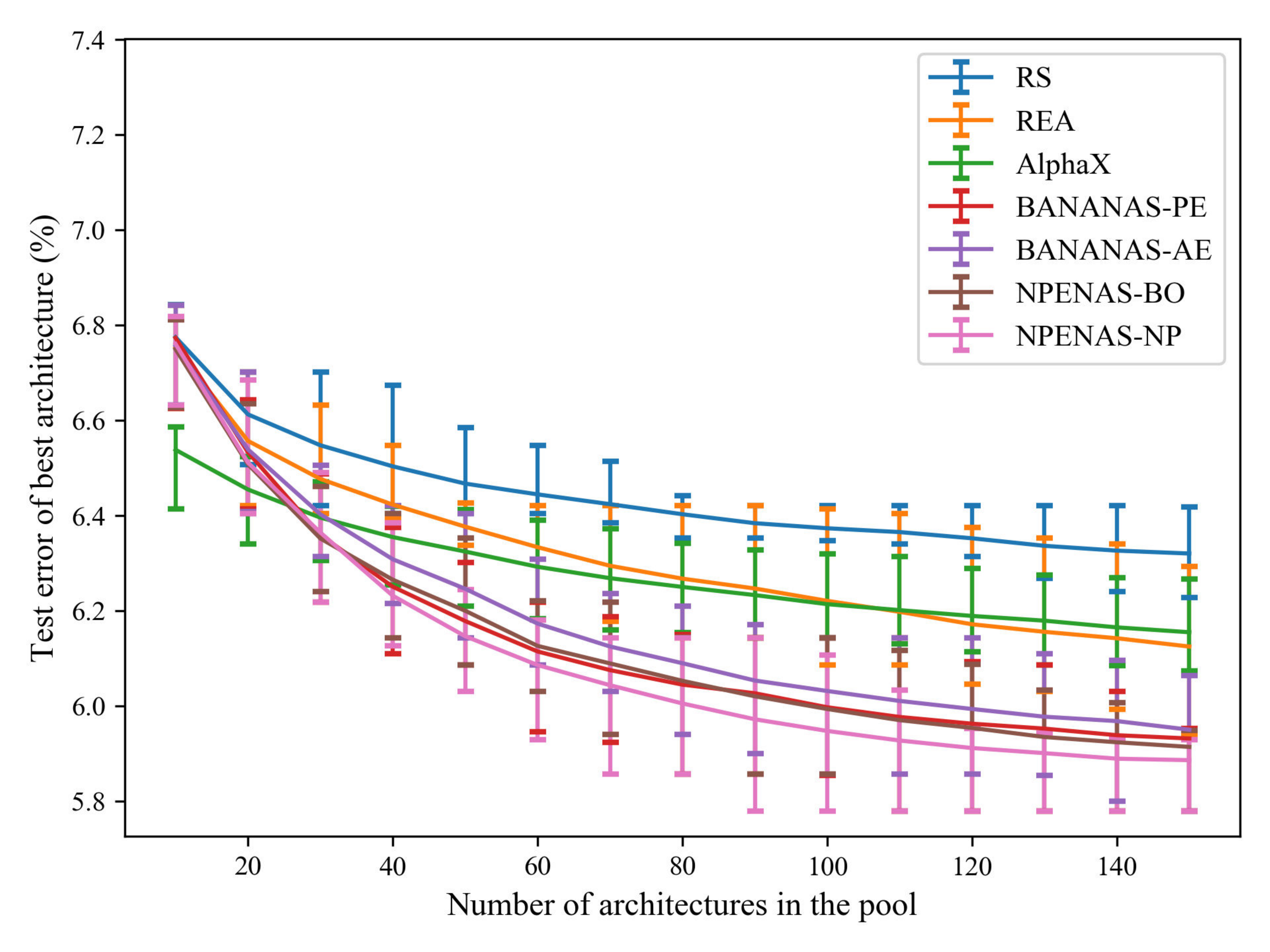}
            \caption{}
            \label{sub_fig:61_nasbench_101_default}
        \end{subfigure}
        \begin{subfigure}{0.48\textwidth}
            \includegraphics[width=0.98\linewidth, height=6.3cm]{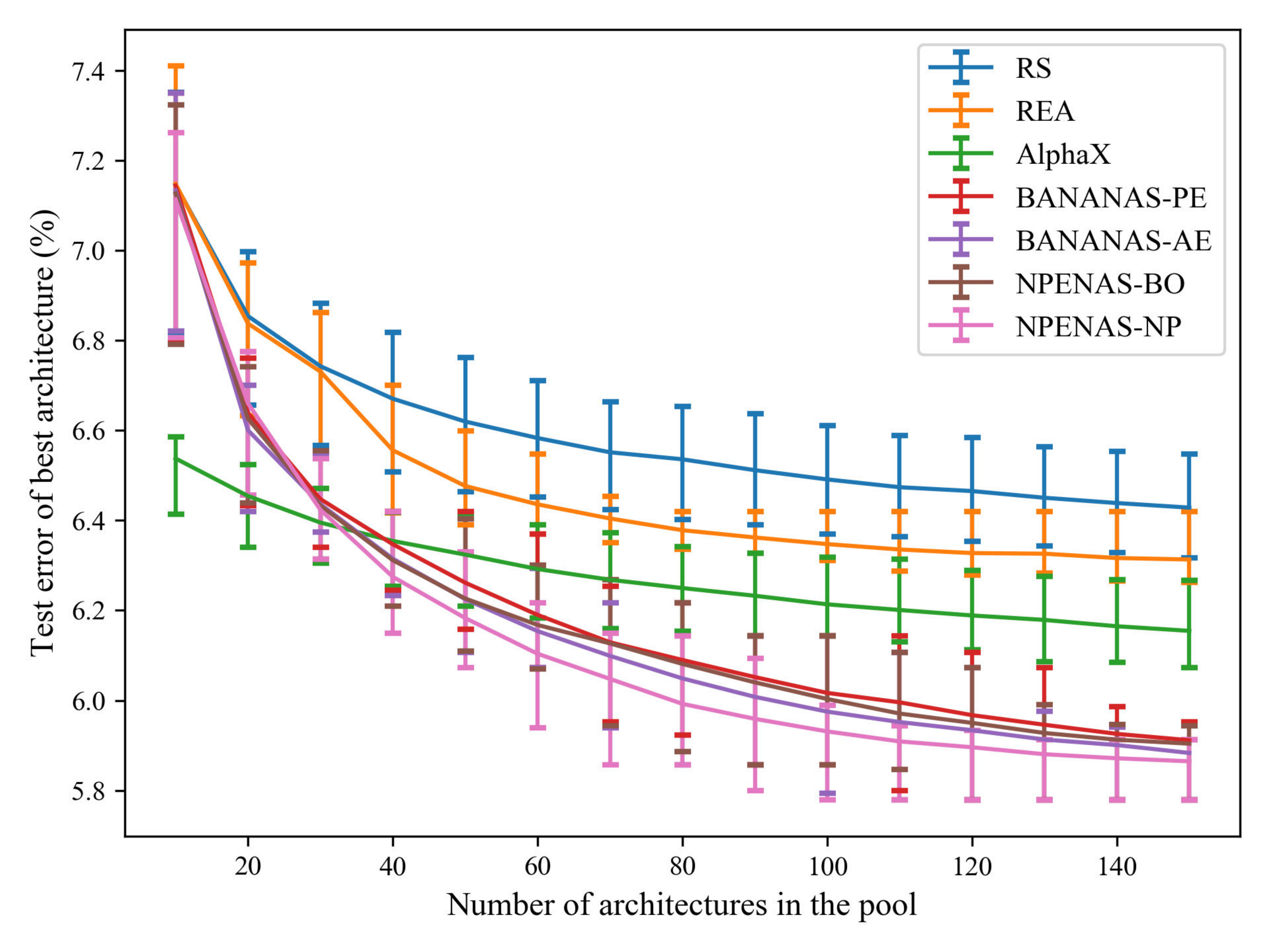}
            \caption{}
            \label{sub_fig:61_nasbench_101_new}
        \end{subfigure}
        \caption{Performance of different NAS algorithms on the NASBench-101 benchmark with (a) the default sampling method and (b) the new sampling method. The error bars represent the $30\%$ to $70\%$ percentile range.}
        \label{fig:6_nas_bench_101_comparisiton}
    \end{center}
\end{figure}

The performance of different algorithms on the NASBench-101 benchmark is shown in Fig. \ref{fig:6_nas_bench_101_comparisiton}. It can be seen that regardless of the sampling method used, the proposed NPENAS-NP outperforms all the other algorithms, and NPENAS-BO is comparable with BANANAS. 
Using the new sampling method and a budget of 150 queries, NPENAS-NP finds a neural architecture with a mean test error of $5.86 \%$, which is very close to the \textit{ORACLE} baseline \cite{Wen2019NeuralPF} $5.77\%$. 
The best neural architecture searched by NPENAS-BO achieves a mean test error of $5.9 \%$, which is slightly better than that of the baseline algorithm BANANAS. The new sampling method brings performance gains to all algorithms using a neural predictor. When changing from the default sampling method to the new sampling method, the mean test error of NPENAS-BO enhances from $ 5.91 \% $ to $ 5.9 \%$, NPENAS-NP from $ 5.89 \% $ to $ 5.86 \% $, BANANAS-PE from $5.93 \% $ to $ 5.91 \%$, and BANANAS-AE from $ 5.94 \% $ to $ 5.88 \% $. 

From Fig. \ref{sub_fig:61_nasbench_101_new}, we can see that BANANAS-AE performs better than BANANAS-PE when using the new sampling method. The result is consistent with the aforementioned findings from Table \ref{table:1} and, again, contradicted with the results in \cite{White2019BANANASBO}. Since predictor performance is critical for NAS algorithms, the increased predictor performance due to the new sampling method will naturally lead to improved NAS algorithm performance. This result reaffirms the need for the new sampling method.

\begin{figure}[ht!]
    \begin{center}
        \includegraphics[width=0.98\linewidth, height=6.3cm]{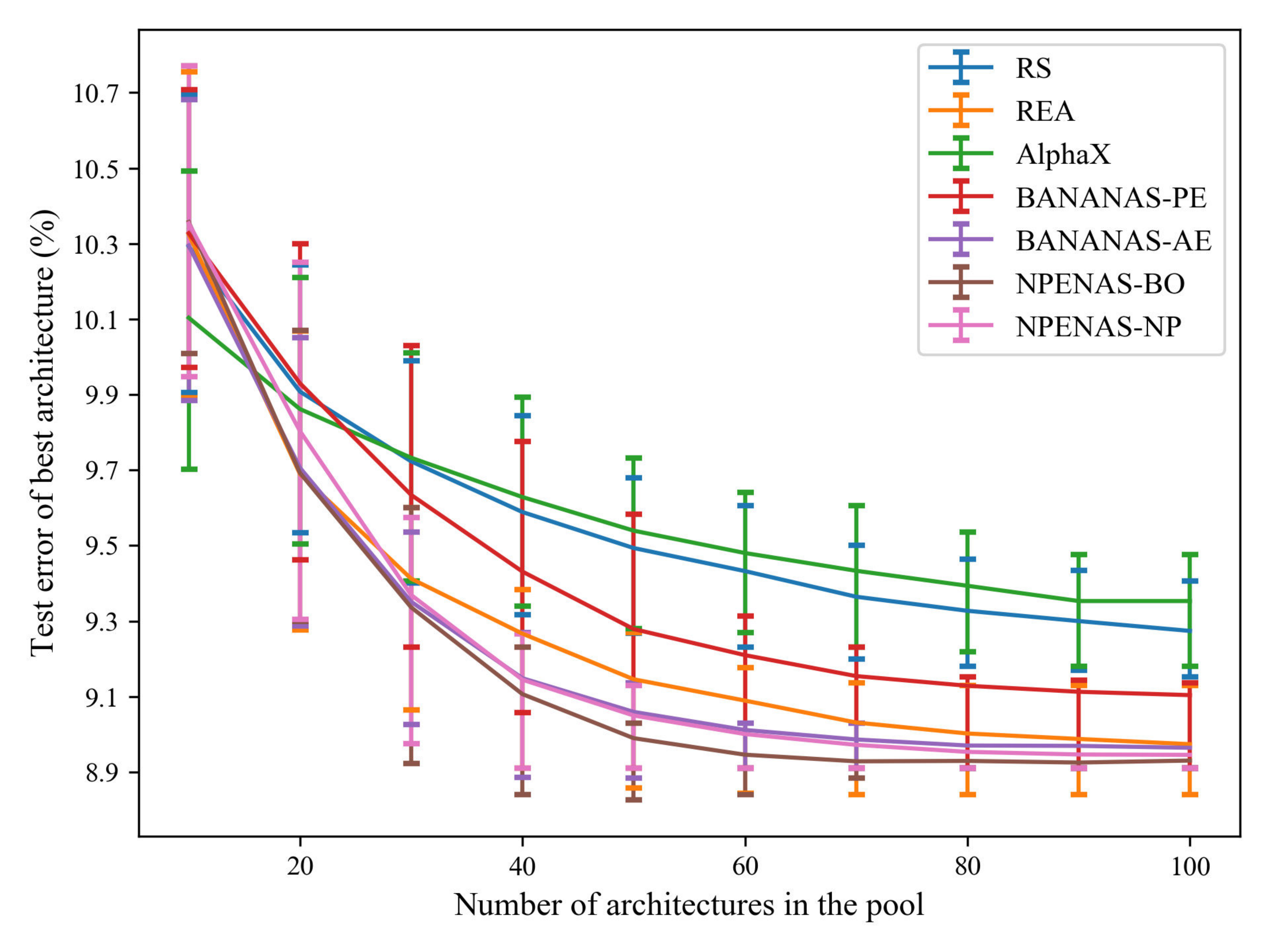}
        \caption{Performance of different NAS algorithms on the NASBench-201 benchmark.}
        \label{nas_bench_201_benchmark}
    \end{center}
\end{figure}

The comparison results on the NASBench-201 benchmark are shown in Fig. \ref{nas_bench_201_benchmark}. The performance of our proposed methods NPENAS-BO and NPENAS-NP achieve outstanding results compared with other algorithms. NPENAS-BO with a budget of $100$ queries achieves a mean test error of $8.93\%$, which is close to the best performance $8.92 \% $. NPENAS-NP with $100$ queries achieves a mean test error of $8.95 \%$, which is better than the strong baseline BANANAS\cite{White2019BANANASBO}.

\subsection{Open Domain Search}
In the case of open domain search, we utilize the DARTS \cite{Liu2018DARTSDA} search space to compare algorithms. 
The search budgets of NPENAS-BO and NPENAS-NP are set at 150 and 100, respectively. Both algorithms employ the same search setting as DARTS \cite{Liu2018DARTSDA}. During the search, a small macro network with eight cells is trained on CIFAR-10 for $50$ epochs with batch size $64$. The CIFAR-10 training dataset is divided into two parts, each containing $25$k images. One part is used for training and the other part for validation. The test images of the CIFAR-10 dataset are not used during the architecture search. Like DARTS, we use momentum SGD with initial learning rate 0.025, momentum 0.9, and weight decay $3 \times 10 ^ {-4}$ to train the macro network. A cosine learning rate schedule \cite{Loshchilov2017SGDRSG} without restart is adopted to annealed down the learning rate to zero. Training enhancements like cutout \cite{Devries2017ImprovedRO}, path dropout, and auxiliary loss are not used during the architecture search. We record the validation accuracies of the sampled architectures at each training epoch. The average validation accuracy is calculated based on the highest validation accuracy and the last five validation accuracies. After the search is completed, only the architecture with the best average validation accuracy is selected to evaluate. 

The evaluation setting of searched architecture is the same as DARTS \cite{Liu2018DARTSDA}. Based on the searched normal cell and reduction cell, a network of $20$ cells with $36$ initial channels is constructed. The network is randomly initialized and trained for 600 epochs with batch size 96. Training enhancements like cutout \cite{Devries2017ImprovedRO}, path dropout, and auxiliary loss are used during architecture evaluation. It takes around $1.5$ GPU days on an Nvidia RTX 2080Ti GPU. The network is trained five times independently with different seeds, and the mean and standard deviation of the test error on CIFAR-10 are reported. 

A comparison of the performance of our NEPNAS method with existing NAS algorithms on the DARTS search space is summarized in Table \ref{table:4}. The search speed and performance of our proposed NPENAS-BO and NPENAS-NP outperform most of the existing NAS algorithms. NPENAS-NP costs 1.8 GPU days.Its speed is 6.5 times faster than the baseline algorithm BANANAS and is comparable with some gradient-based methods, such as DARTS (first-order) 1.5 GPU days. The best architecture found by NPENAS-NP achieves the state-of-the-art test error of $2.44 \%$. NPENAS-BO is 4.7 times faster than the baseline algorithm BANANAS, and the best-searched architecture achieves a test error of $2.52 \%$. It is worth pointing out that our methods achieve state-of-the-art performance with a much smaller search budget compared to other algorithms (100 for NPEANS-NP and 150 for NPENAS-BO). The searched normal cells and reduction cells by NPENAS-BO and NPENAS-NP are visualized in Fig. \ref{fig:10} and Fig. \ref{fig:11}, respectively. 

In order to verify the stability of our proposed algorithms, we carry out four independent experiments on NPENAS-BO and NPENAS-NP, and the results are presented in Appendix \ref{ap:multiple_run_sp}. Our methods can produce stable results over multiple runs. 

\begin{table*}[ht!]
    \normalsize
    \begin{center}
    \caption{Performance comparison of NAS algorithms on DARTS search space}
    \label{table:4}
    \begin{threeparttable}[b]
    \begin{tabular}{ccccccc}
        \hline
        Model  & Params (M)  & Err(\%) Avg & Err(\%) Best & \thead{No. of \\ samples evaluated} & GPU days \\ \hline
        Random search WS \cite{Li2019RandomSA}  &  4.3  &  2.85 $\pm$ 0.08 & 2.71 & -- & 2.7 \\
        NASNet-A \cite{Zoph2017LearningTA}  &  3.3  &  -- & 2.65 & 20000 & 1800 \\
        AmoebaNet-B  \cite{Real2018RegularizedEF}  & 2.8 & 2.55 $\pm$ 0.05  & --  & 27000 & 3150 \\ 
        PNAS \cite{Liu2017ProgressiveNA}  & 3.2 & 3.41 $\pm$ 0.09  & -- & 1160 & 225 \\ 
        ENAS \cite{enas} & 4.6 & -- & 2.89  & -- & 0.45 \\ 
        NAONet \cite{Luo2018NeuralAO}  & 10.6 & -- & 3.18  & 1000 & 200 \\ 
        AlphaX (32 filters) \cite{Wang2019AlphaXEN} & 2.83 & \textbf{2.54 $\pm$ 0.06} & --  & 1000 & 15 \\ 
        ASHA \cite{Li2019RandomSA} &  2.2 &  3.03 $\pm$ 0.13 & 2.85 & 700 & 9 \\
        BANANAS \cite{White2019BANANASBO}  & 3.6$^\dagger$ &  2.64 $\pm$ 0.05 & 2.57 & 100 & 11.8 \\
        GATES \cite{DBLP:journals/corr/abs-2004-01899}  & 4.1 &  -- & 2.58 & 800 & -- \\  
        DARTS (first order) \cite{Liu2018DARTSDA} &  3.3 &  3.00 $\pm$ 0.14 & -- & -- & 1.5 \\
        DARTS (second order) \cite{Liu2018DARTSDA} &  3.3 &  2.76 $\pm$ 0.09 & -- & -- & 4 \\
        SNAS  \cite{Xie2019SNASSN}  & 2.8 & 2.85 $\pm$ 0.02  & -- & -- & 1.5 \\ 
        P-DARTS \cite{Chen2019ProgressiveDA}  & 3.4 & --  & 2.50  & -- & 0.3 \\ 
        BayesNAS \cite{Zhou2019BayesNASAB}  & 3.4 &  2.81 $\pm$ 0.04 & -- & -- & 0.2 \\ 
        PC-DARTS \cite{xu2020pcdarts}  & 3.6 &  2.57 $\pm$ 0.07 & -- & -- & \textbf{0.1} \\ \hline
        NPENAS-BO (ours) & 4.0 & 2.64 $\pm$ 0.08 & 2.52 & 150 & 2.5 \\
        NPENAS-NP (ours) & 3.5 & \textbf{2.54 $\pm$ 0.10} & \textbf{2.44} & 100 & 1.8 \\ \hline
    \end{tabular}
    \begin{tablenotes}
        \item[$\dagger$] {\footnotesize Number of the model parameters is calculated using the genotype provided by the authors.}
    \end{tablenotes}
    \end{threeparttable}
\end{center}
\end{table*}

\section{Conclusion}
In this paper, we propose a neural predictor guided evolutionary algorithm to enhance the exploration ability of EA for NAS (NPENAS) and present two variants of it (NPENAS-BO and NPENAS-NP). NPENAS-BO employs an acquisition function as a neural predictor to guided the evolutionary algorithm to explore the search space. The acquisition function is defined from our designed graph-based uncertainty estimation network. NPENAS-NP employs a graph-based neural predictor to improve the exploration ability of the evolutionary algorithm for NAS. We also analyze the drawbacks of the default architecture sampling method and propose a new sampling method. The proposed NPENAS methods are validated on three NAS search spaces. Extensive experimental results demonstrate that NPENAS achieves state-of-the-art performance. 

How to design an effective architecture representation method that can preserve the structural properties of architectures in the latent space is one of the important future works for NAS algorithms using a neural predictor. Since NPENAS-BO performs better on NASBench-201 while NPENAS-NP better on the other two search spaces, detailed analysis of the applicability of NPENAS-BO and NPENAS-NP constitutes worthy topics of further study. Extending NPENAS-BO and NPENAS-NP to other tasks beyond image classification like semantic segmentation, object detection, and neural machine translation is another meaningful future work. 

\appendices
\section{Structure of the Graph-Based uncertainty Estimation Network} \label{table_detail_app_pp1}
The architecture details of the graph-based uncertainty estimation network in Fig. \ref{fig:npenas_ea}d is presented in Table \ref{table:appendices_pd_1}.

\begin{table}
    \caption{Structure of the graph-based uncertainty estimation network}
\label{table:appendices_pd_1}
\centering
\begin{tabular}{ccc}
    \hline
    Layer & Input dimensional  & Output dimensional  \\ \hline
    GIN-CELU-BN & -- & 32 \\ 
    GIN-CELU-BN & 32 & 32 \\ 
    GIN-CELU-BN & 32 & 32 \\ 
    GMP & 32 & 32 \\ 
    FC-CELU &  32 & 16  \\ 
    Dropout &   16 & 16  \\ \hline
    FC-CELU ($\mu$) &  16  &  1    \\ 
    FC-CELU ($\sigma$)&  16  &  1    \\ \hline 
\end{tabular}
\end{table}

\section{Structure of the Graph-Based neural predictor} \label{table_detail_app_pp2}
The architecture details of the graph-based neural predictor in Fig. \ref{fig:npenas_ea}e is presented in Table \ref{table:appendices_pd_2}.

\begin{table}
    \caption{Structure of the graph-based neural predictor}
\label{table:appendices_pd_2}
\centering
\begin{tabular}{ccc}
    \hline
    Layer & Input dimensional  & Output dimensional  \\ \hline
    GIN-ReLU-BN & -- & 32 \\ 
    GIN-ReLU-BN & 32 & 32 \\ 
    GIN-ReLU-BN & 32 & 32 \\ 
    GMP & 32 & 32 \\ 
    FC-ReLU &  32 & 16  \\ 
    Dropout &  16 & 16  \\ 
    FC-ReLU &  16  &  1    \\ \hline 
\end{tabular}
\end{table}

\section{Illustration of Searched Architecture}
\begin{figure}[ht!]
    \begin{center}
        \begin{subfigure}{0.48\textwidth}
            \includegraphics[width=1.0\linewidth, height=4cm]{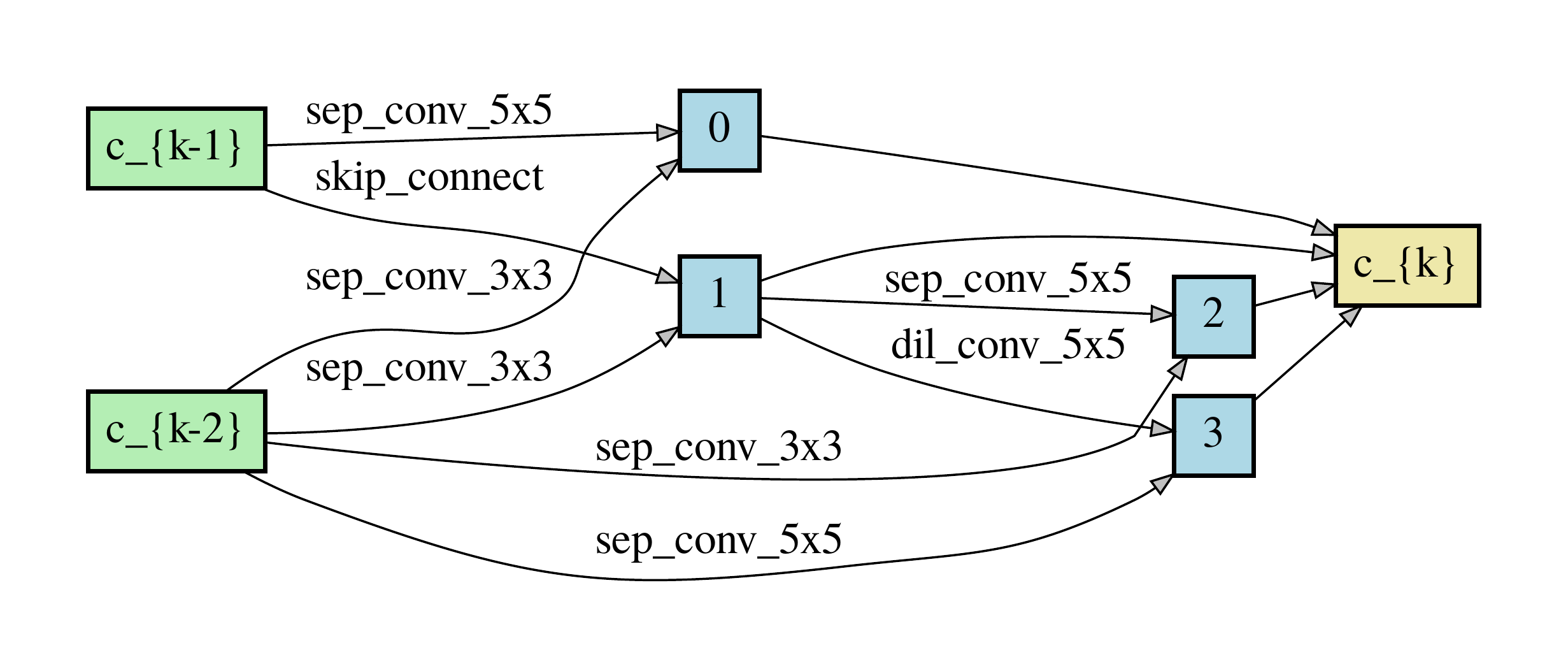}
            \caption{}
            \label{fig:10_61}
        \end{subfigure}
        \begin{subfigure}{0.48\textwidth}
            \includegraphics[width=1.0\linewidth, height=4cm]{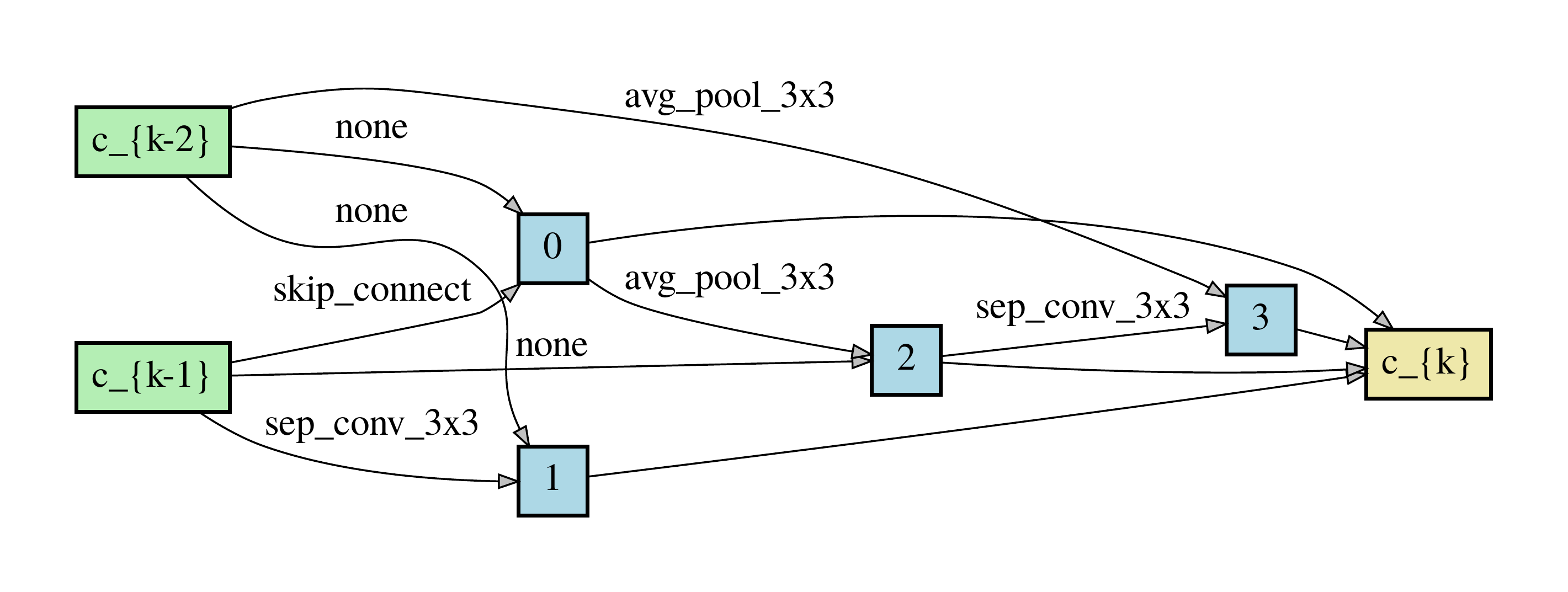}
            \caption{}
            \label{fig:10_62}
        \end{subfigure}
        \caption{(a) The normal cell and (b) reduction cell searched by NPENAS-BO on DARTS.}
        \label{fig:10}
    \end{center}
\end{figure}

\begin{figure}[ht!]
    \begin{center}
        \begin{subfigure}{0.48\textwidth}
            \includegraphics[width=1.0\linewidth, height=4cm]{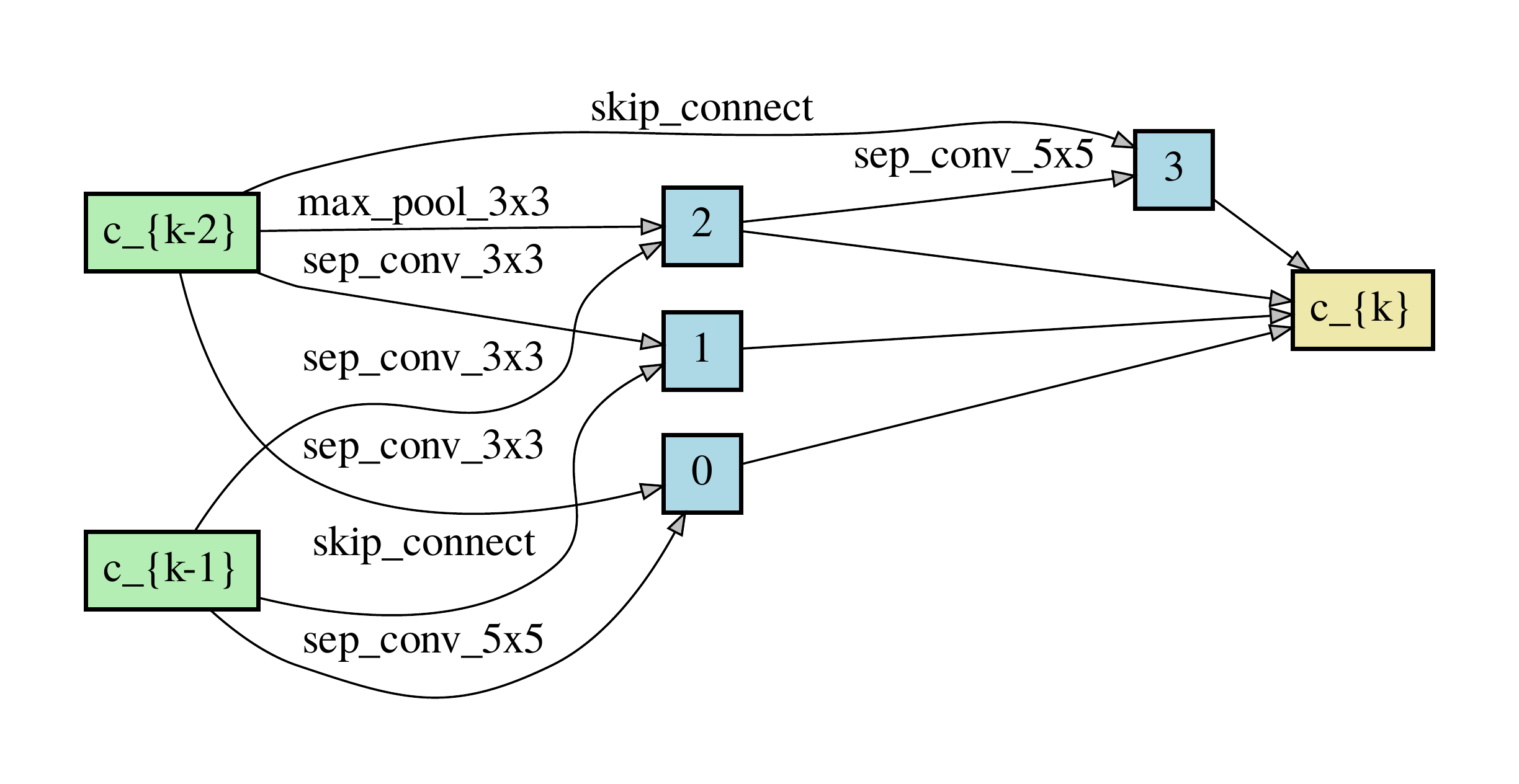}
            \caption{}
            \label{fig:11_61}
        \end{subfigure}
        \begin{subfigure}{0.49\textwidth}
            \includegraphics[width=1.0\linewidth, height=4cm]{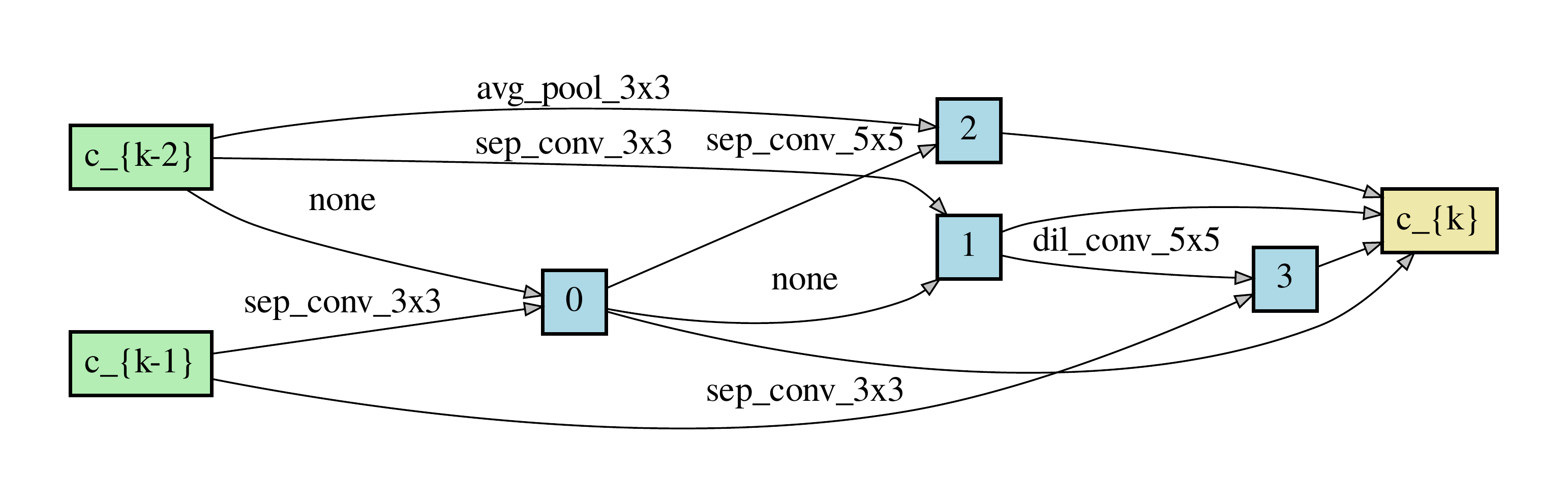}
            \caption{}
            \label{fig:11_62}
        \end{subfigure}
        \caption{(a) The normal cell and (b) reduction cell searched by NPENAS-NP on DARTS.}
        \label{fig:11}
    \end{center}
\end{figure}

The normal cell and reduction cell found by NPENAS-BO and NPENAS-NP on DARTS are illustrated in Fig. \ref{fig:10} and Fig. \ref{fig:11} respectively.

\section{Repeated Search Experiments} \label{ap:multiple_run_sp}
NAS algorithms have large randomness. In order to illustrate the stability of our proposed NPENAS-BO and NPENAS-NP, we conduct four independent search experiments with different seeds and GPU devices. The searched architecture is evaluated five times with different random seeds. From the results in Table \ref{table:a1} and Table \ref{table:a2}, we can find our proposed NPENAS-BO and NPENAS-NP methods are rather stable. 

\begin{table}
    \scriptsize
    \caption{Repeated Search Results of NPENAS-BO}
    \label{table:a1}
    \centering
    \begin{tabular}{ccccc}
        \hline
        \multirow{2}{*}{\thead{Evaluation \\ Seed}} &
        \multicolumn{4}{c}{NPENAS-BO Err ($\%$) Top-1}  \\
        \cmidrule(lr){2-5}
         & Experiment 1 & Experiment 2 & Experiment 3  & Experiment 4 \\ \hline
        0 & 2.72 & 2.83 & 2.70  & 2.52 \\
        1  & 2.72 & 2.64 & 2.80  & 2.61 \\
        2 & 2.78 & 2.65 & 2.71  & 2.63 \\
        3 & 2.75 & 2.65 & 2.54 & 2.75 \\
        4 & 2.98 & 2.59 & 2.68  & 2.71  \\ \hline
        Avg   & 2.79 $\pm$ 0.1 & 2.67 $\pm$ 0.08 & 2.69 $\pm$ 0.08  & \textbf{2.64 $\pm$ 0.08} \\
        Best   & 2.72 & 2.59 & 2.54  & \textbf{2.52} \\ 
        Model Size  & 2.39 M  & 2.7 M & 3.59 M  & 4.0 M \\ \hline
        GPU  & 2080Ti   &  2080Ti  &  Titan Xp  &  Titan Xp \\ \hline
    \end{tabular}
\end{table}

\begin{table}
    \small
    \scriptsize
    \caption{Repeated Search Results of NPENAS-NP}
    \label{table:a2}
    \centering
    \begin{tabular}{ccccc}
        \hline
        \multirow{2}{*}{\thead{Evaluation \\ Seed}}
        &
        \multicolumn{4}{c}{NPENAS-NP Err ($\%$) Top-1}  \\
        \cmidrule(lr){2-5}
         & Experiment 1 & Experiment 2 & Experiment 3  & Experiment 4 \\ \hline
        0 & 2.69 &  2.79 & 2.52 & 2.73 \\
        1  & 2.44 & 2.73 & 2.51 & 2.38 \\
        2 & 2.6 & 2.68 & 2.66 & 2.77  \\
        3 & 2.44  & 2.65 & 2.65 & 2.59 \\
        4 & 2.51 & 2.76 & 2.48  & 2.67 \\  \hline
        Avg   & \textbf{2.54 $\pm$ 0.1} & 2.72 $\pm$ 0.05 & 2.56 $\pm$ 0.08  & 2.63 $\pm$ 0.14 \\
        Best   & 2.44  & 2.65 & 2.48  & \textbf{2.38} \\ 
        Model Size  & 3.53 M  & 3.33 M & 3.26 M  & 3.98 M \\ \hline
        GPU  & Titan Xp &  2080Ti  &  Titan V &  2080Ti   \\ \hline
    \end{tabular}
\end{table}

\ifCLASSOPTIONcaptionsoff
  \newpage
\fi

\bibliographystyle{IEEEtran}
\bibliography{reference_db}

\begin{thebibliography}{10}
\providecommand{\url}[1]{#1}
\csname url@samestyle\endcsname
\providecommand{\newblock}{\relax}
\providecommand{\bibinfo}[2]{#2}
\providecommand{\BIBentrySTDinterwordspacing}{\spaceskip=0pt\relax}
\providecommand{\BIBentryALTinterwordstretchfactor}{4}
\providecommand{\BIBentryALTinterwordspacing}{\spaceskip=\fontdimen2\font plus
\BIBentryALTinterwordstretchfactor\fontdimen3\font minus
  \fontdimen4\font\relax}
\providecommand{\BIBforeignlanguage}[2]{{%
\expandafter\ifx\csname l@#1\endcsname\relax
\typeout{** WARNING: IEEEtran.bst: No hyphenation pattern has been}%
\typeout{** loaded for the language `#1'. Using the pattern for}%
\typeout{** the default language instead.}%
\else
\language=\csname l@#1\endcsname
\fi
#2}}
\providecommand{\BIBdecl}{\relax}
\BIBdecl

\bibitem{Ren2020ACS}
P.~Ren, Y.~Xiao, X.~Chang, P.~Huang, Z.~Li, X.~Chen, and X.~Wang, ``A
  comprehensive survey of neural architecture search: Challenges and
  solutions,'' \emph{ArXiv}, vol. abs/2006.02903, 2020.

\bibitem{Ying2019NASBench101TR}
C.~Ying, A.~Klein, E.~Christiansen, E.~Real, K.~Murphy, and F.~Hutter,
  ``{NAS}-bench-101: Towards reproducible neural architecture search,'' in
  \emph{Proceedings of the 36th International Conference on Machine Learning},
  ser. Proceedings of Machine Learning Research, K.~Chaudhuri and
  R.~Salakhutdinov, Eds., vol.~97.\hskip 1em plus 0.5em minus 0.4em\relax PMLR,
  09--15 Jun 2019, pp. 7105--7114.

\bibitem{kandasamy2018neural}
K.~Kandasamy, W.~Neiswanger, J.~Schneider, B.~Poczos, and E.~P. Xing, ``Neural
  architecture search with bayesian optimisation and optimal transport,'' in
  \emph{Advances in Neural Information Processing Systems}, 2018, pp.
  2016--2025.

\bibitem{White2019BANANASBO}
C.~White, W.~Neiswanger, and Y.~Savani, ``Bananas: Bayesian optimization with
  neural architectures for neural architecture search,'' \emph{ArXiv}, vol.
  abs/1910.11858, 2019.

\bibitem{Real2017LargeScaleEO}
E.~Real, S.~Moore, A.~Selle, S.~Saxena, Y.~L. Suematsu, J.~Tan, Q.~V. Le, and
  A.~Kurakin, ``Large-scale evolution of image classifiers,'' in \emph{ICML},
  2017.

\bibitem{Real2018RegularizedEF}
E.~Real, A.~Aggarwal, Y.~Huang, and Q.~V. Le, ``Regularized evolution for image
  classifier architecture search,'' in \emph{AAAI}, 2019.

\bibitem{9075201}
Y.~{Sun}, B.~{Xue}, M.~{Zhang}, G.~G. {Yen}, and J.~{Lv}, ``Automatically
  designing cnn architectures using the genetic algorithm for image
  classification,'' \emph{IEEE Transactions on Cybernetics}, pp. 1--15, 2020.

\bibitem{8742788}
Y.~{Sun}, B.~{Xue}, M.~{Zhang}, and G.~G. {Yen}, ``Completely automated cnn
  architecture design based on blocks,'' \emph{IEEE Transactions on Neural
  Networks and Learning Systems}, vol.~31, no.~4, pp. 1242--1254, 2020.

\bibitem{8712430}
------, ``Evolving deep convolutional neural networks for image
  classification,'' \emph{IEEE Transactions on Evolutionary Computation},
  vol.~24, no.~2, pp. 394--407, 2020.

\bibitem{Kingma2014AdamAM}
D.~P. Kingma and J.~Ba, ``Adam: A method for stochastic optimization,''
  \emph{CoRR}, vol. abs/1412.6980, 2014.

\bibitem{Dong2020NASBench201ET}
X.~Dong and Y.~Yang, ``Nas-bench-201: Extending the scope of reproducible
  neural architecture search,'' in \emph{8th International Conference on
  Learning Representations, {ICLR} 2020, Addis Ababa, Ethiopia, April 26-30,
  2020}.\hskip 1em plus 0.5em minus 0.4em\relax OpenReview.net, 2020.

\bibitem{Wen2019NeuralPF}
W.~Wen, H.~Liu, H.~Li, Y.~Chen, G.~Bender, and P.~Kindermans, ``Neural
  predictor for neural architecture search,'' \emph{CoRR}, vol. abs/1912.00848,
  2019.

\bibitem{Liu2018DARTSDA}
H.~Liu, K.~Simonyan, and Y.~Yang, ``{DARTS}: Differentiable architecture
  search,'' in \emph{International Conference on Learning Representations},
  2019.

\bibitem{Wang2019AlphaXEN}
L.~Wang, Y.~Zhao, Y.~Jinnai, Y.~Tian, and R.~Fonseca, ``Neural architecture
  search using deep neural networks and monte carlo tree search,'' in \emph{The
  Thirty-Fourth {AAAI} Conference on Artificial Intelligence, {AAAI} 2020, The
  Thirty-Second Innovative Applications of Artificial Intelligence Conference,
  {IAAI} 2020, The Tenth {AAAI} Symposium on Educational Advances in Artificial
  Intelligence, {EAAI} 2020, New York, NY, USA, February 7-12, 2020}.\hskip 1em
  plus 0.5em minus 0.4em\relax {AAAI} Press, 2020, pp. 9983--9991.

\bibitem{DBLP:journals/corr/abs-2004-01899}
X.~Ning, Y.~Zheng, T.~Zhao, Y.~Wang, and H.~Yang, ``A generic graph-based
  neural architecture encoding scheme for predictor-based {NAS},'' in
  \emph{ECCV}, 2020.

\bibitem{Zhou2019BayesNASAB}
H.~Zhou, M.~Yang, J.~Wang, and W.~Pan, ``Bayesnas: A bayesian approach for
  neural architecture search,'' in \emph{ICML}, 2019.

\bibitem{DBLP:conf/iclr/BakerGRN18}
B.~Baker, O.~Gupta, R.~Raskar, and N.~Naik, ``Accelerating neural architecture
  search using performance prediction,'' in \emph{6th International Conference
  on Learning Representations, {ICLR} 2018, Vancouver, BC, Canada, April 30 -
  May 3, 2018, Workshop Track Proceedings}.\hskip 1em plus 0.5em minus
  0.4em\relax OpenReview.net, 2018.

\bibitem{zoph2016neural}
B.~Zoph and Q.~V. Le, ``Neural architecture search with reinforcement
  learning,'' in \emph{International Conference on Learning Representations
  (ICLR)}, 2017.

\bibitem{Kipf2016SemiSupervisedCW}
T.~Kipf and M.~Welling, ``Semi-supervised classification with graph
  convolutional networks,'' \emph{ArXiv}, vol. abs/1609.02907, 2016.

\bibitem{DBLP:journals/spm/ShumanNFOV13}
D.~I. Shuman, S.~K. Narang, P.~Frossard, A.~Ortega, and P.~Vandergheynst, ``The
  emerging field of signal processing on graphs: Extending high-dimensional
  data analysis to networks and other irregular domains,'' \emph{{IEEE} Signal
  Process. Mag.}, vol.~30, no.~3, pp. 83--98, 2013.

\bibitem{DBLP:conf/icml/WuSZFYW19}
F.~Wu, A.~H.~S. Jr., T.~Zhang, C.~Fifty, T.~Yu, and K.~Q. Weinberger,
  ``Simplifying graph convolutional networks,'' in \emph{Proceedings of the
  36th International Conference on Machine Learning, {ICML} 2019, 9-15 June
  2019, Long Beach, California, {USA}}, ser. Proceedings of Machine Learning
  Research, K.~Chaudhuri and R.~Salakhutdinov, Eds., vol.~97.\hskip 1em plus
  0.5em minus 0.4em\relax {PMLR}, 2019, pp. 6861--6871.

\bibitem{Xu2018HowPA}
K.~Xu, W.~Hu, J.~Leskovec, and S.~Jegelka, ``How powerful are graph neural
  networks?'' in \emph{7th International Conference on Learning
  Representations, {ICLR} 2019, New Orleans, LA, USA, May 6-9, 2019}, 2019.

\bibitem{Gilmer2017NeuralMP}
J.~Gilmer, S.~S. Schoenholz, P.~F. Riley, O.~Vinyals, and G.~E. Dahl, ``Neural
  message passing for quantum chemistry,'' \emph{ArXiv}, vol. abs/1704.01212,
  2017.

\bibitem{enas}
H.~Pham, M.~Y. Guan, B.~Zoph, Q.~V. Le, and J.~Dean, ``Efficient neural
  architecture search via parameter sharing,'' in \emph{ICML}, 2018.

\bibitem{Zoph2017LearningTA}
B.~Zoph, V.~Vasudevan, J.~Shlens, and Q.~V. Le, ``Learning transferable
  architectures for scalable image recognition,'' in \emph{2018 IEEE/CVF
  Conference on Computer Vision and Pattern Recognition}, 2017, pp. 8697--8710.

\bibitem{Liu2017ProgressiveNA}
C.~Liu, B.~Zoph, M.~Neumann, J.~Shlens, W.~Hua, L.-J. Li, L.~Fei-Fei,
  A.~Yuille, J.~Huang, and K.~Murphy, ``Progressive neural architecture
  search,'' in \emph{Computer Vision -- ECCV 2018}, V.~Ferrari, M.~Hebert,
  C.~Sminchisescu, and Y.~Weiss, Eds.\hskip 1em plus 0.5em minus 0.4em\relax
  Springer International Publishing, 2018, pp. 19--35.

\bibitem{Xie2019SNASSN}
S.~Xie, H.~Zheng, C.~Liu, and L.~Lin, ``{SNAS}: stochastic neural architecture
  search,'' in \emph{International Conference on Learning Representations},
  2019.

\bibitem{Chen2019ProgressiveDA}
X.~Chen, L.~Xie, J.~Wu, and Q.~Tian, ``Progressive differentiable architecture
  search: Bridging the depth gap between search and evaluation,'' \emph{2019
  IEEE/CVF International Conference on Computer Vision (ICCV)}, pp. 1294--1303,
  2019.

\bibitem{xu2020pcdarts}
Y.~Xu, L.~Xie, X.~Zhang, X.~Chen, G.-J. Qi, Q.~Tian, and H.~Xiong,
  ``{PC}-{DARTS}: Partial channel connections for memory-efficient architecture
  search,'' in \emph{International Conference on Learning Representations},
  2020.

\bibitem{Elsken2018NeuralAS}
T.~Elsken, J.~H. Metzen, and F.~Hutter, ``Neural architecture search: A
  survey,'' \emph{Journal of Machine Learning research}, vol.~20, pp.
  55:1--55:21, 2018.

\bibitem{Li2019RandomSA}
L.~Li and A.~Talwalkar, ``Random search and reproducibility for neural
  architecture search,'' \emph{ArXiv}, vol. abs/1902.07638, 2019.

\bibitem{Sciuto2019EvaluatingTS}
K.~Yu, C.~Sciuto, M.~Jaggi, C.~Musat, and M.~Salzmann, ``Evaluating the search
  phase of neural architecture search,'' in \emph{International Conference on
  Learning Representations}, 2020.

\bibitem{Shahriari2016TakingTH}
B.~Shahriari, K.~Swersky, Z.~Wang, R.~P. Adams, and N.~de~Freitas, ``Taking the
  human out of the loop: A review of bayesian optimization,'' \emph{Proceedings
  of the IEEE}, vol. 104, pp. 148--175, 2016.

\bibitem{Frazier2018ATO}
P.~I. Frazier, ``A tutorial on bayesian optimization,'' \emph{CoRR}, vol.
  abs/1807.02811, 2018.

\bibitem{Russo2017ATO}
D.~Russo, B.~V. Roy, A.~Kazerouni, and I.~Osband, ``A tutorial on thompson
  sampling,'' \emph{Foundations and Trends in Machine Learning}, vol.~11, pp.
  1--96, 2017.

\bibitem{DBLP:books/lib/Bishop07}
C.~M. Bishop, \emph{Pattern recognition and machine learning, 5th Edition},
  ser. Information science and statistics.\hskip 1em plus 0.5em minus
  0.4em\relax Springer, 2007.

\bibitem{Paszke2019PyTorchAI}
A.~Paszke, S.~Gross, F.~Massa, A.~Lerer, J.~Bradbury, G.~Chanan, T.~Killeen,
  Z.~Lin, N.~Gimelshein, L.~Antiga, A.~Desmaison, A.~K{\"o}pf, E.~Yang,
  Z.~DeVito, M.~Raison, A.~Tejani, S.~Chilamkurthy, B.~Steiner, L.~Fang,
  J.~Bai, and S.~Chintala, ``Pytorch: An imperative style, high-performance
  deep learning library,'' in \emph{NeurIPS}, 2019.

\bibitem{FeyLenssen2019}
M.~Fey and J.~E. Lenssen, ``Fast graph representation learning with {PyTorch
  Geometric},'' in \emph{ICLR Workshop on Representation Learning on Graphs and
  Manifolds}, 2019.

\bibitem{Npenascode}
\BIBentryALTinterwordspacing
C.~Wei. (2020) Code for {NPENAS}. [Online]. Available:
  \url{https://github.com/auroua/NPENASv1}
\BIBentrySTDinterwordspacing

\bibitem{Krizhevsky09learningmultiple}
A.~Krizhevsky, ``Learning multiple layers of features from tiny images,'' Tech.
  Rep., 2009.

\bibitem{Barron2017ContinuouslyDE}
J.~T. Barron, ``Continuously differentiable exponential linear units,''
  \emph{ArXiv}, vol. abs/1704.07483, 2017.

\bibitem{Ioffe2015BatchNA}
S.~Ioffe and C.~Szegedy, ``Batch normalization: Accelerating deep network
  training by reducing internal covariate shift,'' \emph{ArXiv}, vol.
  abs/1502.03167, 2015.

\bibitem{Krizhevsky2017ImageNetCW}
A.~Krizhevsky, I.~Sutskever, and G.~E. Hinton, ``Imagenet classification with
  deep convolutional neural networks,'' in \emph{CACM}, 2017.

\bibitem{Loshchilov2017SGDRSG}
I.~Loshchilov and F.~Hutter, ``Sgdr: Stochastic gradient descent with warm
  restarts,'' in \emph{ICLR}, 2017.

\bibitem{Devries2017ImprovedRO}
T.~Devries and G.~W. Taylor, ``Improved regularization of convolutional neural
  networks with cutout,'' \emph{ArXiv}, vol. abs/1708.04552, 2017.

\bibitem{Luo2018NeuralAO}
R.~Luo, F.~Tian, T.~Qin, and T.-Y. Liu, ``Neural architecture optimization,''
  in \emph{NeurIPS}, 2018.

\end{thebibliography}

\end{document}